\documentclass[journal]{IEEEtran}
\usepackage{ifpdf}
\ifCLASSOPTIONcompsoc
  \usepackage[nocompress]{cite}
\else
  \usepackage{cite}
\fi
\usepackage{nomencl}
\usepackage{amssymb}
\usepackage[flushleft]{threeparttable}
\usepackage{url}
 \usepackage{eufrak}
\usepackage{longtable}
\usepackage{booktabs}
\usepackage{enumitem}
\usepackage[ruled,vlined]{algorithm2e}
\usepackage{mathtools}
\SetKwComment{Comment}{/* }{ */}
\SetKwRepeat{Do}{do}{while}
\usepackage{breqn}

\usepackage{graphicx}
\usepackage{wrapfig}
\usepackage{lscape}
\usepackage{multirow}
\usepackage{rotating}
\usepackage{amsmath}

\usepackage{xcolor}
\usepackage{amsthm}
\theoremstyle{theorem} 

  \usepackage[caption=false,font=footnotesize]{subfig}

\usepackage{epstopdf}
\usepackage[nolist]{acronym}
\usepackage{mathtools}
\theoremstyle{definition}

\definecolor{charcoal}{RGB}{54, 69, 79}
\definecolor{steel-gray}{RGB}{113, 121, 126}

\begin{document}
\newacro{dcnn}[DCNN]{deep convolutional neural network}
\newacro{nas}[NAS]{neural architecture search}
\newacro{dc}[DC]{deep clustering}
\newacro{cl}[CL]{contrastive learning}
\newacro{ncl}[NCL]{non-contrastive learning}
\newacro{rl}[RL]{representation learning}
\newacro{pip}[PIP]{positive instances proximity}
\newacro{cdr}[CDR]{cluster dispersion regularizer}
\newacro{mm}[MM]{majorize-minimization}
\title{Enhancing Clustering Representations with Positive Proximity and Cluster Dispersion Learning}

\author{Abhishek~Kumar,
        and~Dong-Gyu~Lee\IEEEauthorrefmark{2}
\thanks{A. Kumar, and D.G. Lee are with the Department of Artificial Intelligence, Kyungpook National University, Daegu, Republic of Korea - 41566}
\thanks{\IEEEauthorrefmark{2} Corresponding author (e-mail: dglee@knu.ac.kr).}
\thanks{Manuscript received XXXXX, 2023; revised August XXXXX, 2023.}}

%
%

\markboth{Journal of \LaTeX\ Class Files}%
{Shell \MakeLowercase{\textit{et al.}}: Bare Advanced Demo of IEEEtran.cls for IEEE Computer Society Journals}
%



\maketitle 

\begin{abstract}
Contemporary deep clustering approaches often rely on either contrastive or non-contrastive techniques to acquire effective representations for clustering tasks. Contrastive methods leverage negative pairs to achieve homogenous representations but can introduce class collision issues, potentially compromising clustering performance. On the contrary, non-contrastive techniques prevent class collisions but may produce non-uniform representations that lead to clustering collapse. In this work, we propose a novel end-to-end deep clustering approach named PIPCDR, designed to harness the strengths of both approaches while mitigating their limitations. PIPCDR incorporates a positive instance proximity loss and a cluster dispersion regularizer. The positive instance proximity loss ensures alignment between augmented views of instances and their sampled neighbors, enhancing within-cluster compactness by selecting genuinely positive pairs within the embedding space. Meanwhile, the cluster dispersion regularizer maximizes inter-cluster distances while minimizing within-cluster compactness, promoting uniformity in the learned representations. PIPCDR excels in producing well-separated clusters, generating uniform representations, avoiding class collision issues, and enhancing within-cluster compactness. We extensively validate the effectiveness of PIPCDR within an end-to-end Majorize-Minimization framework, demonstrating its competitive performance on moderate-scale clustering benchmark datasets and establishing new state-of-the-art results on large-scale datasets.
\end{abstract}



\IEEEdisplaynontitleabstractindextext

%
\IEEEpeerreviewmaketitle

\ifCLASSOPTIONcompsoc
\IEEEraisesectionheading{\section{Introduction}\label{sec:introduction}}
\else
\section{Introduction}
\label{sec:introduction}
\fi

\IEEEPARstart{C}{lustering} represents a foundational task within machine learning and data mining, aiming to classify data into clusters devoid of annotation effectively. The overarching objective is to ensure that data points within each cluster belong to the same class or exhibit shared semantic characteristics. Beyond facilitating general \ac{rl}~\cite{caron2018deep,caron2020unsupervised}, clustering plays a pivotal role in a wide range of real-world applications, spanning from face recognition~\cite{shen2021structure,wang2022ada,shin2023local} and medical analysis~\cite{ali2018neutrosophic,thanh2017neutrosophic,chang2018deep} to gene sequencing~\cite{kiselev2019challenges,lahnemann2020eleven}.

In recent years, the research area of clustering has perceived a substantial shift towards advancing \ac{dc} methods~\cite{ghasedi2017deep,li2020autoencoder}. Unlike their predecessors~\cite{chen2009spectral,liu2017sparse,liu2016multiple,tang2019kernel} that primarily emphasized the design of similarity metrics and clustering strategies, these modern approaches leverage deep learning techniques to extract highly representative features. This paradigm shift has proven promising, as it overcomes the limitations associated with shallow models~\cite{ren2022deep,zhou2022comprehensive}. Early \ac{dc} methods~\cite{caron2018deep,peng2016deep,xie2016unsupervised,yang2016joint} adopt an iterative approach, where \ac{rl} and clustering mutually reinforce each other in a bootstrapping manner. The primary driving force behind the progress in \ac{dc} lies in self-supervised \ac{rl} techniques, encompassing both \ac{cl}~\cite{he2020momentum,chen2020simple} and \ac{ncl} approaches~\cite{grill2020bootstrap,chen2021exploring,huang2022learning}.

\begin{figure*}
    \centering
    \includegraphics[width=0.8\linewidth]{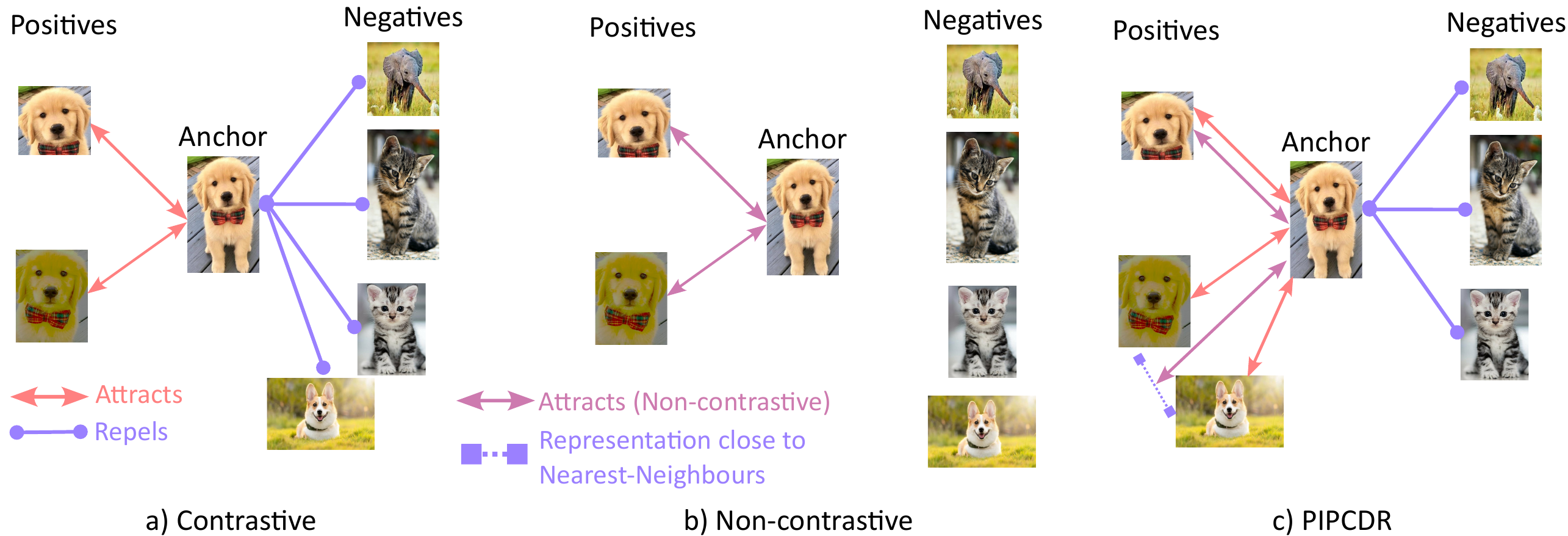}
    \caption{A compelling illustration of the class collision issues arises in \ac{cl}, accompanied by the problem of clustering collapse encountered in \ac{ncl}. \textit{a)} In a typical \ac{cl}-based approach, the augmented instances belonging to different entities are conventionally treated as negative pairs. However, considering the semantic similarities among a few examples, such as instances of other dogs, it would be more reasonable to treat them as positive pairs. \textit{b)} In the \ac{ncl}-based methodologies, a potential quandary can arise wherein the embedding vectors of instances of different classes start converging and collapsing onto a singular point. This convergence phenomenon can detrimentally impact the discernibility and distinctiveness of individual data points. \textit{c)} To tackle these significant challenges, the proposed method integrates contrastive and non-contrastive \ac{rl}, effectively mitigating class collision issues, enhancing clustering stability through uniform representations, and improving cluster separation and within-cluster compactness.}
    \label{fig:1_new}
\end{figure*}

Significantly, prevailing \ac{dc} models depend on contrastive \ac{rl}, commonly termed as contrastive-based models~\cite{wang2021unsupervised,van2020scan}. Specifically, these methods are often built upon frameworks such as MoCo~\cite{he2020momentum} or SimCLR~\cite{chen2020simple}, necessitating the adoption of clustering-friendly losses~\cite{wang2021unsupervised,li2020prototypical,li2021contrastive,tao2021clustering,tsai2021mice} or an additional pre-training stage to foster the development of more discriminative representations~\cite{van2020scan,niu2022spice}. While contrastive-based models have shown great potential in achieving effective clustering results, their effectiveness often hinges on the availability of a significant amount of negative examples to facilitate the learning of uniform representations within an embedding space where instances are distinctly separated. However, this reliance on a multitude of negative pairs may inadvertently give rise to a critical challenge known as the \textbf{class collision issue}. In this scenario, instances belonging to the same semantic class are erroneously treated as negative pairs, leading to their forced separation within the embedding space. This unintended separation can have a detrimental impact on downstream clustering tasks~\cite{arora2019theoretical}. A substitute approach to address this concern is to decompose the conventional contrastive loss into two distinct terms~\cite{wang2020understanding}: \textit{1)} an \textit{alignment} term aimed at enhancing the proximity of positive pairs, and \textit{2)} a \textit{uniformity} term intended to promote the uniform distribution of instances on a unit hypersphere by separating them from negative pairs. However, it should be noted that the \textit{uniformity} term can potentially lead to a class collision issue~\cite{arora2019theoretical}, as the negative pairs constructed may not genuinely represent instances that are dissimilar, as shown in Fig. \ref{fig:1_new}.a.

\Ac{ncl} stands apart from contrastive-based methodologies through its exclusive integration of the alignment element. Non-contrastive-based methods effectively circumvent the class collision issue by omitting the use of negative pairs. Due to the absence of the uniformity term in non-contrastive loss, the learning process does not inherently prioritize the attainment of uniform representations~\cite{wang2020understanding,zhang2022does}. Consequently, this can lead to the \textbf{representation collapse} of downstream clustering, where a large majority of samples are assigned to only a few clusters, as shown in Fig. \ref{fig:1_new}.b. This occurrence becomes further pronounced when integrated with extra clustering losses implemented within the advanced \ac{dc} approaches~\cite{li2020prototypical,li2021contrastive}.

We introduce an innovative end-to-end \ac{dc} method called PIPCDR to tackle the challenges of class collision and representation collapse. This method integrates a novel loss with two terms: \textit{positive instances proximity} loss and \textit{cluster dispersion regularizer}. Firstly, to enhance the compactness within clusters, we introduce the \textit{positive instances proximity} loss. This loss facilitates the alignment of one augmented view of an instance with the randomly sampled neighbors of another view of the same instance, bringing them in close proximity to each other. In contrast to traditional alignment methods that primarily focus on aligning two augmented views, our proposed PIP approach takes a more comprehensive perspective by considering the influence of neighboring samples within the embedding space. By doing so, it effectively promotes within-cluster compactness, a crucial aspect of clustering quality. Secondly, recognizing that instances belonging to different clusters are indeed negative pairs, we introduce \ac{cl} as a means to learn uniform representations and promote well-separated clusters. In this approach, two augmented views of instances from the same cluster are considered positive pairs, while instances from different clusters serve as negative pairs. This formulation gives rise to the proposed \textit{cluster dispersion regularizer}, which maximizes the between-cluster distance and facilitates learning uniform representations for distinct clusters. Furthermore, we employ a \ac{mm} framework in the training steps of PIPCDR. The first M-step involves estimating the pseudo-labels of each instance using spherical $k$-means. Subsequently, we focus on minimizing the proposed losses in the second M-step. This approach ensures a proficient and systematic optimization of PIPCDR.

We can summarize the key contributions of our work as follows:
\begin{itemize}
    \item This work introduces an end-to-end \ac{dc} technique, named PIPCDR, that combines contrastive and non-contrastive \ac{rl}. This method effectively addresses class collision, enhances clustering stability with uniform representations, and improves both cluster separation and within-cluster compactness.
    \item In this paper, we propose the \ac{pip} loss, which expands instance alignment by incorporating neighboring positive examples. This inclusion significantly enhances the within-cluster compactness, leading to improved clustering results.
    \item We present a novel approach called the \ac{cdr}, which aligns clusters by minimizing intra-cluster distance and maximizing inter-cluster dispersion on the unit hypersphere. The CDR significantly enhances clustering performance by promoting both within-cluster compactness and between-cluster separation.
    \item Through optimizing PIPCDR within a \ac{mm} framework, our extensive experiments on benchmark datasets demonstrate the superior performance of PIPCDR. It outperforms state-of-the-art approaches by a considerable gap, particularly on moderate- and large-scale datasets. These results confirm the efficacy and competitiveness of our proposed model.
\end{itemize}

The subsequent sections of this paper are structured in the following ways. Section 2 provides a concise overview of the pertinent literature on self-supervised learning and \ac{dc}. In Section 3, we introduce our proposed PIP and CDR techniques, along with the comprehensive PIPCDR framework. Experimental results are presented and analyzed in Section 4, with specific emphasis on the substantiation of our motivation in Section 4.3. Section 5 discusses the connections to prior research efforts, culminating in a comprehensive summary and concluding remarks in Section 7.
\section{Related Works}
This section provides a concise summary of the progress made in the fields of self-supervised \ac{rl} and \ac{dc}.
\subsection{Contrastive Learning}
In recent years, the \ac{cl} paradigm has demonstrated its effectiveness in self-supervised \ac{rl}~\cite{chen2020simple,chen2021exploring,he2020momentum,zbontar2021barlow}. Through the initial generation of positive and negative pairs for each instance, a solid foundation is established for subsequent mapping into a dedicated subspace. This mapping endeavor is focused on maximizing the similarities among positive pairs while efficiently minimizing those among the negatives~\cite{hadsell2006dimensionality}. The most commonly employed strategy involves leveraging labels to steer the process of constructing pairs~\cite{khosla2020supervised}. However, in the context of unsupervised learning, it becomes imperative to employ alternative strategies for the construction and optimal utilization of contrastive pairs. For instance, SimCLR~\cite{chen2020simple} methodology adopts an approach of constructing positive and negative pairs by applying augmentations within a mini-batch. On the other hand, MoCo~\cite{he2020momentum} tackles \ac{cl} by transforming it into a dictionary look-up task, accomplished through the creation of a dynamic dictionary utilizing a queue and a moving-averaged encoder. In recent years, the efficacy of these methods has become increasingly evident, showcasing promising results in \ac{rl} as well as downstream tasks~\cite{li2021contrastive}. Contrastive \ac{rl} necessitates a substantial amount of negative pairs to attain fine-grained differentiation among instances within an embedding space that ensures clear separation among all instances~\cite{he2020momentum,chen2020simple}. The construction of negative pairs typically demands a substantial batch size~\cite{chen2020simple}, a memory queue~\cite{he2020momentum}, or a memory bank~\cite{wu2018unsupervised}, resulting in additional computational overhead. However, this approach can lead to a class collision issue~\cite{saunshi2019theoretical}, where semantically similar examples are erroneously moved apart as they might be treated as negative pairs. Nevertheless, despite efforts to mitigate the class collision issue, it still persists as a challenge in contrastive \ac{rl}. Researchers have made several attempts to tackle this problem~\cite{khosla2020supervised,chuang2020debiased}.

\subsection{Non-contrastive Learning}
Recent studies have shown that positive examples are enough for \ac{rl}, leading to the emergence of non-contrastive methods~\cite{grill2020bootstrap}. BYOL~\cite{grill2020bootstrap} and SimSiam~\cite{chen2021exploring} introduce an online predictor as a replacement for negative pairs, effectively preventing the network from converging to trivial solutions. As an alternative approach, AdCo~\cite{hu2021adco} adopts an adversarial learning framework to learn negative samples directly. However, it is worth noting that some attempts to address the class collision issue using \ac{ncl} have yielded inferior results~\cite{leelearning,regatti2021consensus}. These methods frequently encounter challenges related to the collapse of downstream clustering, predominantly attributed to the non-uniform representations they generate~\cite{huang2022learning}. Recently, ProPos~~\cite{huang2022learning} addresses the collapse of downstream clustering by maximizing between-cluster distance, resulting in uniform representations and improved clustering stability. It also incorporates positive sampling alignment to enhance within-cluster compactness, leading to superior clustering performance.

\subsection{Deep Clustering}
The achievement of good clustering relies on both effective clustering strategies and discriminative features. Thanks to the powerful representability of deep neural networks, \ac{dc} methods have gained significant attention in recent years~\cite{asano2019self,caron2018deep,guo2017improved,li2022twin,peng2022xai}. These methods have been greatly enhanced through the utilization of self-supervised \ac{rl}~\cite{leelearning,huang2022learning,li2021contrastive,shen2021you}. The majority of \ac{dc} methods leverage \ac{cl} to obtain discriminative representations, which are then used to enhance downstream clustering tasks~\cite{van2020scan,niu2022spice,li2022twin} or jointly optimize \ac{rl} and clustering~\cite{li2020prototypical,tao2021clustering,tsai2021mice,shen2021you,guo2022hcsc}. For instance, the SCAN~\cite{van2020scan} leverages the pre-trained SimCLR model to generate confident pseudo-labels. On the other hand, IDFD~\cite{tao2021clustering} introduces a combined approach of instance discrimination and feature decorrelation. 
WCL~\cite{zheng2021weakly} and GCC~\cite{zhong2021graph} utilize a strategy where neighbor instances from a graph are selected as pseudo-positive examples for calculating contrastive loss. Nonetheless, these methods encounter an ongoing challenge of class collision, as the representation of positive pairs may not be accurately reflected by the selected instances. In line with the framework of \ac{cl}, these approaches necessitate many negative instances to establish uniform and consistent representations. Unfortunately, this necessity unavoidably gives rise to the emergence of class collision.

\section{Proposed Method}
The primary objective of \ac{dc} is to learn image representations while simultaneously performing clustering tasks effectively. In this context, our proposed method, PIPCDR, introduces innovative techniques such as \ac{pip} loss and \ac{cdr}, which are comprehensively detailed in this section.
\subsection{Preliminary}
\subsubsection{Contrastive Learning}
\Ac{cl} methods leverage the InfoNCE loss~\cite{oord2018representation} to facilitate instance-wise discrimination~\cite{wu2018unsupervised}. Formally, let us consider an instance $x_i$, its augmented versions $x_i^a$ and $x_i^b$ obtained through random data augmentation, and a set of $2(N-1)$ negative samples extracted from other instances of the dataset with $N$ instances. The main aim of \ac{cl} is to train an embedding function $f(\cdot)$ that effectively maps the instance $x$ onto a unit hypersphere. This is achieved by minimizing the InfoNCE loss, which quantifies the alignment between positive pairs $(x_i^a, x_i^b)$ and the discrepancy between negative pairs $(x_i^a, x_j^c)$, where $j \in \{1, 2, ..., N\}\backslash\{i\}$ and $c \in \{a,b\}$. We aim to enhance the discriminative power of the learned representations and encourage instances belonging to the same class to be close in the embedding space by optimizing the embedding function $f(\cdot)$, while pushing instances from different classes apart. The InfoNCE loss for $x_i^a$ can be defined in the following way:
\begin{equation}\label{eqn_1}
    \mathcal{L}(x_i^a) = -\log \frac{e^{\frac{f(x_i^a)^Tf(x_i^b)}{\tau}}}{e^{\frac{f(x_i^a)^Tf(x_i^b)}{\tau}}+\sum_{j=1,j\neq i}^N \sum_{c\in\{a,b\}}e^{ \frac{f(x_i^a)^Tf(x_j^c)}{\tau}}},
\end{equation}
where the representation $f(x)$ is $\ell_2$ normalized to lie on a unit hypersphere, ensuring that the embeddings have unit length. Additionally, the parameter $\tau$ is introduced to regulate the concentration level of the representations. The sharpness of the learned similarity scores is modulated by adjusting the value of $\tau$. Higher values of $\tau$ result in a softer distribution over the negative examples, promoting exploration during training, while lower values encourage more concentrated representations.

As demonstrated in \cite{wang2020understanding}, the InfoNCE loss shown in Eq. (\ref{eqn_1}) can be effectively decoupled into two distinct terms, resulting in the following approximation.
\begin{equation}\label{eqn_2}
    \mathcal{L}(x_i^a) \approx - \frac{f(x_i^a)^Tf(x_i^b)}{\tau} + \log \sum_{j=1,j\neq i}^N \sum_{c\in\{a,b\}}e^{ \frac{f(x_i^a)^Tf(x_j^c)}{\tau}},
\end{equation}
In this context, the initial term is referred to as instance alignment, while the subsequent term is known as instance uniformity. While the alignment term effectively brings the positive pair closer, it is the uniformity term that plays a crucial role in preventing representation collapse, as it promotes a uniform distribution of negative examples on the unit hypersphere. However, it is important to note that despite its benefits in mitigating downstream clustering collapse, the inclusion of negative examples may introduce the challenge of class collision~\cite{arora2019theoretical}, which can potentially undermine the quality of the representations for clustering purposes.
\subsubsection{Non-contrastive Learning}
Non-contrastive methods, on the other hand, focus solely on optimizing the instance alignment term in Eq. (\ref{eqn_2}) to facilitate greater proximity between the representations of two augmented views closer~\cite{grill2020bootstrap,chen2021exploring}. Typically, these methods employ a predictor network, a target network, and an online network to connect the disparity between the two different augmented perspectives, using stop gradient operations to prevent the collapse of representations. Specifically, when the $\tau$ is set to 0.5, the loss function employed in \cite{grill2020bootstrap,chen2021exploring} can be formulated as:
\begin{equation}\label{eqn_3}
    -2gf(x_i^a)^Tf'(x_i^b) = ||g(f(x_i^a)) -f'(x_i^b) ||_2^2-2,
\end{equation}
where $f(.)$, $f'(.)$, and $g(.)$  represent the online network, target network, and predictor network, respectively. The terms $g(f(x_i^a))$ and $f'(x_i^b)$ are $\ell_2$-normalized. Non-contrastive methods effectively mitigate the issue of class collision by avoiding the use of negative pairs. However, the absence of uniformity in \ac{ncl} methods often gives rise to non-uniform representations, which can lead to the collapse of downstream clustering algorithms. This lack of uniformity poses a significant challenge in \ac{dc} tasks, as it introduces instability and hinders the overall effectiveness of the clustering process.

\subsection{Positive Instances Proximity Loss}
Contrastive-based \ac{dc} heavily relies on negative examples to achieve uniform representations, albeit at the expense of encountering the class collision issue~\cite{arora2019theoretical}, adversely affecting within-cluster compactness. In contrast, non-contrastive-based methods circumvent the problem of class collisions by focusing solely on minimizing the alignment between positive instances. However, the traditional formulation of instance alignment, shown in Eq. (\ref{eqn_3}), predominantly emphasizes the proximity between augmented views while neglecting the incorporation of semantic class information. Consequently, this approach fails to promote the desired within-cluster compactness that is crucial for effective \ac{dc}~\cite{li2021contrastive,li2020prototypical}, as it operates solely at the instance level without accounting for semantic class distinctions.

In order to address the limitations of conventional instance alignment and mitigate the class collision issue, we present a novel approach that focuses on enhancing within-cluster compactness by promoting alignment between neighboring examples instead of pursuing uniformity. These selected neighboring instances, considered to be precise positive pairs, are facilitated to align with another augmented view, thereby fostering a more cohesive representation. Our motivation stems from the understanding that while it may not be feasible to guarantee the authenticity of negative pairs created from the dataset, it is reasonable to consider that neighboring examples surrounding one view are indeed positive in relation to other augmented views and share the same semantic characteristic. With this in mind, we introduce a \ac{pip}  approach that builds upon the instance alignment in Eq. (\ref{eqn_3}) by incorporating neighboring samples. This extension aims to enhance within-cluster compactness and improve the overall clustering performance. By incorporating the nearest neighbors of the instances in Eq. (\ref{eqn_3}), the proposed PIP method is expressed as follows:
\begin{equation}\label{eqn_4}
    \mathcal{L}_{\mbox{pip}}(x_i^a) = ||g(v_i) - f'(x_i^b)||^2_2,
\end{equation}
where $v_i = f(x_i^a) + \sigma (f(x_{nn_i}^a) - f(x_i^a))$ and 
\begin{equation*}
nn_i= \underset{j}{\mbox{argmin}}~\left(1-f(x_i^a)^Tf(x_j^a)\right),~j\in\{1,\ldots,N\}\backslash \{i\}.
\end{equation*}
In this formulation, when the parameter $\sigma$ is set to 0, the PIP simplifies to Eq. (\ref{eqn_3}). In contrast to the conventional instance alignment used in \ac{ncl}, our proposed PIP in Eq. (\ref{eqn_4}) focuses on encouraging the alignment of nearby instances surrounding one augmented view. These examples encompass various augmented views of the same examples or distinct examples within the same cluster. We aim to enhance the within-cluster compactness of the learned representations by considering these positive pairs. In non-contrastive methods where negative instances are not used, proposed PIP loss ensures that the positive examples surrounding an instance, sampled from the embedding space, belong to the same cluster and form genuine positive pairs. The proposed method effectively addresses the issue of class collision, which is a common challenge in \ac{dc} models based on \ac{cl}, through the optimization of PIP.

\subsection{Cluster Dispersion Regularizer}
A clustering algorithm aims to achieve well-separated clusters. Let us consider a dataset consisting of $K$ classes $\{c_k\}_{k=1}^K$, where $K$ can be a known parameter or predefined in advance. It is natural to construct a contrastive loss that considers multiple positive samples belonging to one cluster and treats the remaining samples from other clusters as negative examples~\cite{khosla2020supervised}.
\begin{equation}\label{eqn_5}
    \mathcal{L'}(x_i^a) = -\log \frac{e^{\frac{f(x_i^a)^Tf(x_i^b) + \sum_{k \in p(i)\backslash \{i\}} \sum_{c \in \{a,b\}} f(x_i^a)^Tf(x_k^c)}{(2|p(i)|-1)\tau}}}{e^{\frac{f(x_i^a)^Tf(x_i^b)}{\tau}}+\sum_{j=1,j\neq i}^N \sum_{c\in\{a,b\}}e^{ \frac{f(x_i^a)^Tf(x_j^c)}{\tau}}}.
\end{equation}
Here, $p(i)$ represents the set of instances belonging to the cluster to which $i$ belongs, i.e., $p(i)=\{j:j\in c_k,~\mbox{where}~i\in c_k\}$.

In the proposed \ac{cdr}, we approximate the loss function $\mathcal{L'}(x_i^a)$ in a manner similar to the formulation presented in Eq. (\ref{eqn_2}) to separate out the alignment and uniformity term.
\begin{equation}\label{eqn_6}
\begin{split}
    \mathcal{L}_{\mbox{cdr}}(x_i^a) = &- \frac{f(x_i^a)^Tf'(x_i^b) + \sum\limits_{k \in p(i)\backslash \{i\}} \sum\limits_{c \in \{a,b\}} f(x_i^a)^Tf'(x_k^c)}{(2|p(i)|-1)\tau} \\ &+ \log \sum_{j=1,j\not\in p(i)}^N \sum_{c\in\{a,b\}}e^{ \frac{f(x_i^a)^Tf'(x_j^c)}{\tau}}.
\end{split}
\end{equation}
During the training phase, the exact  $\{c_k\}_{k=1}^K$ remains unknown in an unsupervised setting, which presents a challenge when optimizing the proposed CDR directly. To tackle this issue, a \ac{mm} framework is employed. The $k$-means clustering algorithm is leveraged to substitute Eq. (\ref{eqn_6}) with a surrogate function in each epoch. We can define a surrogate function for  Eq. (\ref{eqn_6}) as follows:
\begin{equation}\label{eqn_7}
\begin{split}
    \mathcal{L}_{\mbox{cdr}}^{'}(x_i^a) = &- \frac{f(x_i^a)^Tf'(x_i^b) + \sum\limits_{k \in \tilde{p}(i)\backslash \{i\}} \sum\limits_{c \in \{a,b\}} f(x_i^a)^Tf'(x_k^c)}{(2|\tilde{p}(i)|-1)\tau} \\ &+ \log \sum_{j=1,j\not\in \tilde{p}(i)}^N \sum_{c\in\{a,b\}}e^{ \frac{f(x_i^a)^Tf'(x_j^c)}{\tau}},
\end{split}
\end{equation}
where $\tilde{p}(i) = \{j:j\in\tilde{c}_k,~\mbox{where}~i\in \tilde{c}_k\}$. Here, $\{\tilde{c}_k\}_{k=1}^K$ is estimated using $k$-means algorithm. This surrogate function is subsequently minimized to optimize the CDR. The detailed procedures and methodologies pertaining to this process will be elaborated on in the subsequent sections. 

From an intuitive standpoint, our CDR shares similarities with conventional supervised contrastive loss~\cite{khosla2020supervised}. However, the key distinction lies in its application within the self-supervised setting. Remarkably, even in this self-supervised context, CDR effectively mitigates the class collision issue by designating different clusters as unequivocal negative examples for each other's constituents. This unique characteristic renders CDR highly suitable for \ac{dc} tasks. Nevertheless, it is worth noting that the accuracy of the clusters may not meet expectations during the initial stages of training. To ensure an accurate initialization, we adopt a strategy similar to the one outlined in \cite{huang2022learning,li2020prototypical}, whereby CDR is incorporated into the training process only after the completion of the learning rate warmup phase.

The proposed CDR can be decomposed into two distinct terms: cluster alignment and cluster uniformity. The cluster alignment term aims to minimize intra-cluster distances, thus promoting stability in cluster updates. Conversely,  the cluster uniformity term promotes the uniform distribution of clusters on a unit hypersphere, effectively maximizing inter-cluster distances. Our CDR facilitates the creation of well-separated and cohesive clusters by combining these two objectives. In \ac{dc}, non-contrastive methods lack a uniformity term, which hinders the generation of uniform representations, leading to the potential collapse of downstream clustering. To overcome this limitation, the proposed CDR method introduces a novel approach that maximizes the inter-cluster distance among cluster representations. This enables the generation of uniform representations and fosters the formation of distinct and well-separated clusters.

\subsection{Sketch of PIPCDR and its Training}
\begin{figure*}[!ht]
    \centering
    \includegraphics[width=0.7\linewidth]{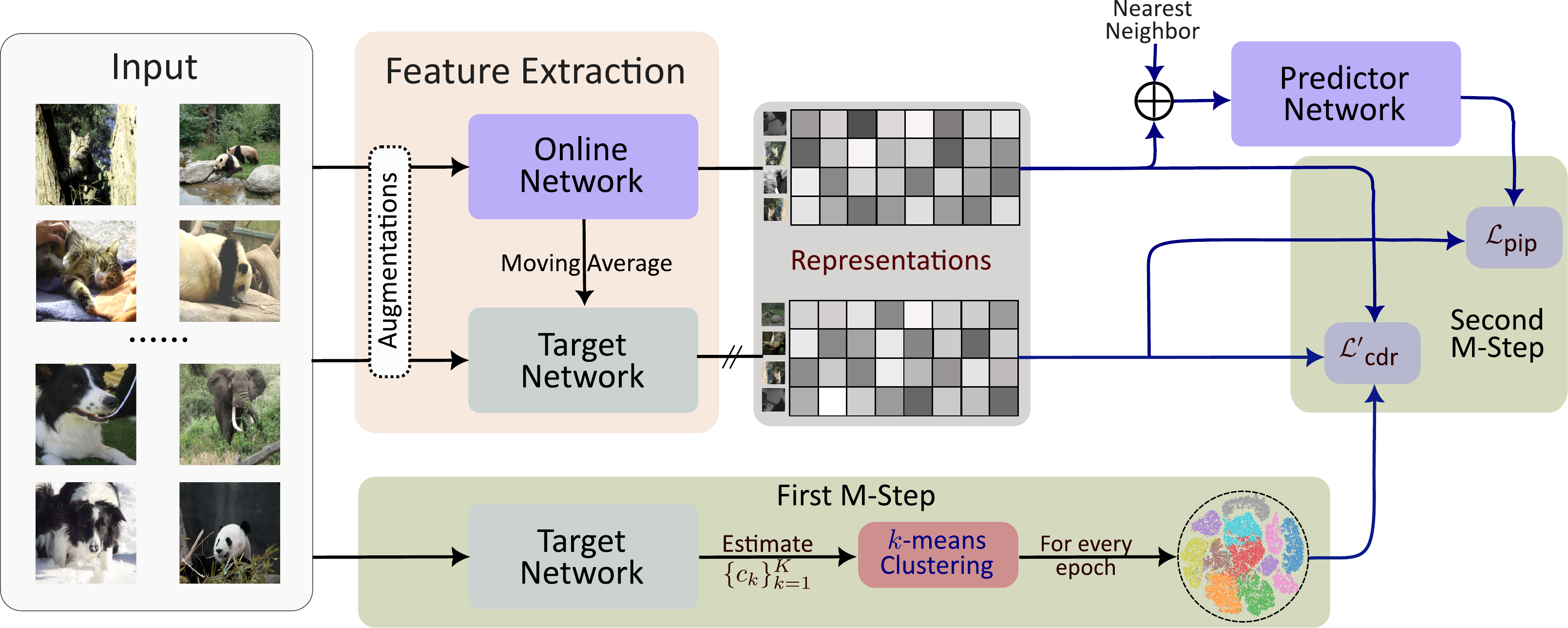}
    \caption{The overall framework of PIPCDR comprises three interconnected networks: a predictor network, a target network, and an online network. The training of PIPCDR is carried out within an MM framework, consisting of two crucial steps: the first M-step and the second M-step. 
    }
    \label{fig:1}
\end{figure*}
We provide an overview of our proposed method, PIPCDR, in Fig. \ref{fig:1}, illustrating the key components and their interactions. To enhance the comprehension of the training process, the optimization of PIPCDR is discussed within a \ac{mm} framework.
\subsubsection{Model Overview}
We construct PIPCDR based on a non-contrastive-based methodology, which shares similarities with BYOL~\cite{grill2020bootstrap}. This framework consists of three networks: a predictor network, a target network, and an online network. Throughout the training process, the parameters of the target network undergo updates using momentum from the parameters of the online network. This update follows the equation:
\begin{equation}
    \theta_{target} = m\theta_{target} + (1-m)\theta_{online},
\end{equation}
where the coefficient $m \in [0, 1)$ represents the momentum coefficient, $\theta_{target}$ and $\theta_{online}$ represent the target and online network parameters, respectively.

\subsubsection{Majorize-Minimization Framework}
The \ac{mm} framework is a widely adopted optimization technique employed to address complex optimization problems that are challenging to solve directly. It revolves around constructing a surrogate function that provides an upper bound on the objective function of interest, thereby facilitating easier optimization. We conduct the optimization of PIPCDR within an MM framework comprising two steps: the first M-step and the second M-step. We provide a detailed explanation of each step below.
\begin{itemize}[labelsep=-2.2cm, listparindent = 5ex, leftmargin = 0cm]
    \item[\textbf{First M-Step:} ] \hspace{2.0cm} In this step, we aim to estimate a surrogate function for the proposed CDR. To achieve this, we apply the spherical $k$-means algorithm to the features extracted from the target network. The choice of the target network is motivated by its stability and ability to yield consistent clusters, which aligns with the principles of BYOL and MoCo. Finally, we construct the surrogate function within each mini-batch using the pseudo-labels, eliminating the requirement for additional memory.

\end{itemize}
\begin{itemize}[labelsep=-2.6cm, listparindent = 5ex, leftmargin = 0cm]
    \item[\textbf{Second M-Step:} ] \hspace{2.4cm} To initiate the second M-step, two different random data augmentations are applied to the same inputs and fed into the target and online networks during training. During the second step, we compute the PIP loss on the representations obtained from both networks. Subsequently, we transform the representation of the online network by leveraging the nearest neighbor within the predictor network to facilitate effective alignment. Then, we utilize the obtained $\{\tilde{c}_k\}_{k=1}^K$ from the first M-step along with the representations obtained from the online and target networks to calculate CDR using Eq. (\ref{eqn_7}). Combining the PIP loss term and the surrogate function of CDR, we obtain our objective function for the second M-step as follows:
\end{itemize}
\begin{equation}\label{eqn_9}
    \mathcal{L} = w\mathcal{L}_{\mbox{pip}} + (1-w)\mathcal{L'}_{\mbox{cdr}},
\end{equation}
where the loss weight parameter $w$ controls the balance between the two loss components. Therefore, the loss function in our method involves only two hyperparameters, namely: $\sigma$ in PIP loss and $w$ in the final loss function. We present the training procedure of our proposed PIPCDR method in Algorithm \ref{algo:1}.

A high-quality clustering approach must exhibit within-cluster compression and well-distinct clusters. To achieve this, our proposed PIPCDR method leverages two key components: CDR and PIP. CDR aims to enhance cluster separation and within-cluster compactness by minimizing intra-cluster distance and maximizing inter-cluster distance. On the other hand, PIP pulls together neighboring examples from the same augmented view and another view, resulting in enhanced compactness within all clusters. Combining these two losses in Eq. (\ref{eqn_9}), PIPCDR aims to effectively improve \ac{dc} by promoting compactness within each cluster and fostering well-separated clusters.  

Despite the need for an additional $k$-means clustering step to acquire cluster pseudo-labels every $r$ epochs, our approach does not incur substantial computational overhead. Through extensive experimentation (as detailed in Section 4.4.1), we observe consistent performance improvements compared to baseline methods, even when employing larger values of $r~(> 1)$. As a result, our method demonstrates both computational efficiency and resilience to cluster pseudo-labels, thereby minimizing the introduction of significant supplementary computational expenses.
\begin{algorithm}
    \DontPrintSemicolon
    \SetKwRepeat{Repeat}{repeat}{until}  
    \SetKwBlock{FStep}{First M-step}{}
    \SetKwBlock{SStep}{Second M-step}{}
    \caption{ Training of PIPCDR}
    \label{algo:1}
    \KwIn{Dataset $\{x_i\}_{i=1}^N$}\;\vspace{-3mm}
    \KwOut{Clusters $\{\tilde{c}_k\}_{k=1}^K$}\;\vspace{-3mm}
    Initialize all training parameters\;
    \Repeat{Reaching maximum Epochs}
        {\FStep{update $\{\tilde{c}_k\}_{k=1}^K$ for $\{x_i\}_{i=1}^N$ using $k$-means\;}
        \SStep{
        Randomly select a mini-batch from $\{x_i\}_{i=1}^N$\;
        \For{each $x_i$ in the batch}
        {
        $\mathcal{L}_{\mbox{pip}} \gets$ Eq. (\ref{eqn_4})\;
        $\mathcal{L'}_{\mbox{cdr}} \gets$ Eq. (\ref{eqn_7})\;
        }
        $\mathcal{L} \gets$ Eq. (\ref{eqn_9})\;
        Update Target Network's parameters using momentum moving average\;
        Update Online and Predictor Networks' parameters using SGD optimizer\;
        }}  
\end{algorithm}

\textit{Remark 1:} In light of the constraints imposed by page limitations, we provide a concise discussion on the relationship between our proposed model and other baseline methods in the supplementary material. In this supplementary section, we aim to offer readers a comprehensive understanding of how our approach compares and contrasts with existing baselines, delving into the nuances and key differentiators that underpin the superiority of our proposed model.

\section{Experiments}
The evaluation of the clustering performance of the proposed PIPCDR is conducted on six publicly available image datasets in this section. A comprehensive and intuitive understanding of the method, a range of qualitative analyses, and ablation studies are performed.
\subsection{Experimental Arrangement}
\subsubsection{Datasets}
We carried out experiments on a comprehensive set of benchmark datasets, which includes CIFAR-10~\cite{krizhevsky2009learning}, CIFAR-20~\cite{krizhevsky2009learning}, ImageNet-Dogs~\cite{chang2017deep}, ImageNet-10~\cite{chang2017deep}, ImageNet-1k~\cite{deng2009imagenet}, and Tiny-ImageNet~\cite{le2015tiny}. We present a concise overview of the dataset characteristics in Table \ref{tab:ex_1}. In the case of CIFAR-20~\cite{krizhevsky2009learning}, the ground truth is established based on the 20 super-classes of CIFAR-100, rather than the 100 fine-grained classes. This paper adheres to the commonly employed experimental settings~\cite{tao2021clustering, tsai2021mice} utilized in \ac{dc} research, which encompass the train-test split, backbone, and, image size. Specifically, we utilize a size of 32 $\times$ 32 for CIFAR-10 and CIFAR-20, and 224 $\times$ 224 for Tiny-ImageNet,  ImageNet-Dogs, ImageNet-10, and ImageNet-1k. For datasets such as CIFAR-10 and CIFAR-20, we choose to utilize the complete dataset, encompassing both the training and testing sets. 
\begin{table}[!ht]
    \centering
    \caption{Brief description of the datasets.}
    \resizebox{\linewidth}{!}{\begin{tabular}{ccccc}\hline
        Dataset & Split & \#Samples & \#Classes & Image Size \\\hline
        CIFAR-10~\cite{krizhevsky2009learning} & Train+Test & 60000  & 10  & 32  \\
        CIFAR-20~\cite{krizhevsky2009learning} & Train+Test &  60000 & 20  & 32  \\
        ImageNet-10~\cite{chang2017deep} & Train  &  13000 & 10  & 224  \\
        ImageNet-Dogs~\cite{chang2017deep}   & Train  &  19500 & 15  & 224 \\
        Tiny-ImageNet~\cite{le2015tiny}   &  Train &  100000 & 200  & 224  \\
        ImageNet-1k~\cite{deng2009imagenet} & Train  &  1281167 & 1000  & 224  \\\hline
    \end{tabular}}
    \label{tab:ex_1}
\end{table}
\subsubsection{ResNet Backbones}
To ensure the integrity of our assessments and to present our primary results on benchmark datasets of moderate scale, ImageNet-10, ImageNet-Dogs, and Tiny-ImageNet, we employ ResNet-34~\cite{he2016deep} as the backbone. For CIFAR-10 and CIFAR-20, we utilize ResNet-18~\cite{he2016deep}, while for ImageNet-1k, we follow the existing literature and utilize ResNet-50~\cite{he2016deep}. Unless otherwise specified, ResNet-18 is used for the remaining experiments. In consideration of the relatively small image sizes of CIFAR-10 and CIFAR-20, as suggested in \cite{chen2020simple}, we modify the architecture by substituting the initial convolution layer with a kernel size of 7 $\times$ 7 and stride of 2, with a convolution layer of kernel size 3 $\times$ 3 and stride of 1. Additionally, we exclude the first max-pooling layer in all experiments conducted on CIFAR-10 and CIFAR-20.

\subsubsection{Implementation Setup}
During the implementation process, we meticulously adhere to the recommended settings of the state-of-the-art models to ensure the reproduction of results, thus enabling fair comparisons. For the training of PIPCDR, we adhere strictly to the methodology outlined in the literature~\cite{tsai2021mice,grill2020bootstrap} and set the number of epochs to 1000. We employ the stochastic gradient descent (SGD) optimizer with a cosine decay learning rate schedule. The learning rate is warmed up over 50 epochs, starting from a base rate of 0.05. It is linearly scaled with the batch size using the formula LearningRate = 0.05 x BatchSize/256. It is important to highlight that the learning rates for the predictor networks of PIPCDR are ten times higher than the learning rate utilized for the feature extractor. This distinction is crucial for achieving satisfactory performance, as discussed in \cite{grill2020bootstrap}. Regarding the remaining hyperparameters of PIPCDR, we set the temperature ($\tau$) to 0.1, the loss weight ($w$) to 0.91, and the sampling noise rate ($\sigma$) to 0.001.

In our approach, we utilize the same data augmentations as SimCLR~\cite{chen2020simple}, incorporating techniques such as ResizedCrop, ColorJitter, Grayscale, and HorizontalFlip. However, we exclude GaussianBlur from our augmentation pipeline when we use small image-sized datasets. In line with the methodology of BYOL~\cite{grill2020bootstrap}, we adhere to the same settings. Our networks employ standard ResNet backbones, while the architectures of the projection and predictor networks are FC-BN-ReLU-FC. For both networks, the projection dimension is set to $256$, and the hidden size is $4096$. These consistent settings ensure comparability and allow for meaningful evaluation and analysis. To compute the loss, we have implemented a symmetric approach by swapping the two data augmentations. This allows us to effectively utilize the augmented data. Additionally, we set the momentum hyperparameter, denoted as $m$ and belonging to the interval $[0, 1)$, to the same value as ~\cite{grill2020bootstrap}, which is $0.996$. This choice ensures consistency and facilitates fair comparisons with the referenced work~\cite{grill2020bootstrap}.
\subsection{Clustering Results}
\begin{table*}[htbp]
  \centering
  \setlength{\tabcolsep}{9.5pt}
  \caption{Comparison of the clustering results in terms of NMI, ACC, and ARI of competing methods on four benchmark datasets: CIFAR-10, CIFAR-20, ImageNet-10, and ImageNet-Dog. The methods are categorized based on their respective training paradigms. Notably, the IDFD, PCL, and ProPos approaches are particularly relevant to our method as they aim to enhance representations specifically for clustering purposes.}
    \begin{tabular}{lcccccccccccc}
    \toprule
    \multirow{2}[3]{*}{\textbf{Methods}} & \multicolumn{3}{c}{\textbf{CIFAR-10}} & \multicolumn{3}{c}{\textbf{CIFAR-20}} & \multicolumn{3}{c}{\textbf{ImageNet-10}} & \multicolumn{3}{c}{\textbf{ImageNet-Dogs}} \\
\cmidrule{2-13}          & NMI & ACC & ARI & NMI & ACC & ARI & NMI & ACC & ARI & NMI & ACC & ARI \\\midrule
    IIC~\cite{ji2019invariant}   & 0.513 & 0.617 & 0.411 &   --    & 0.257 &   --    &   --    &    --   &    --   &   --    &   --    &  --\\
    DCCM~\cite{wu2019deep}  & 0.496 & 0.623 & 0.408 & 0.285 & 0.327 & 0.173 & 0.608 & 0.710 & 0.555 & 0.321 & 0.383 & 0.182 \\
    PICA~\cite{huang2020deep}  & 0.561 & 0.645 & 0.467 & 0.296 & 0.322 & 0.159 & 0.782 & 0.850 & 0.733 & 0.336 & 0.324 & 0.179 \\\midrule
    SCAN~\cite{van2020scan}  & 0.797 & 0.883 & 0.772 & 0.486 & 0.507 & 0.333 &   --    &   --    &   --    &    --   &  --     & -- \\
    NMM~\cite{dang2021nearest}   & 0.748 & 0.843 & 0.709 & 0.484 & 0.477 & 0.316 &  --     &  --     &   --    &  --     &  --     & -- \\\midrule
    CC~\cite{li2021contrastive}    & 0.705 & 0.790 & 0.637 & 0.431 & 0.429 & 0.266 & 0.859 & 0.893 & 0.822 & 0.445 & 0.429 & 0.274 \\
    MiCE~\cite{tsai2021mice}  & 0.737 & 0.835 & 0.698 & 0.436 & 0.440 & 0.280 &    --   &   --    &  --     & 0.423 & 0.439 & 0.286 \\
    GCC~\cite{zhong2021graph}   & 0.764 & 0.856 & 0.728 & 0.472 & 0.472 & 0.305 & 0.842 & 0.901 & 0.822 & 0.490 & 0.526 & 0.362 \\
    TCL~\cite{li2022twin}   & 0.819 & 0.887 & 0.780 & 0.529 & 0.531 & 0.357 & 0.875 & 0.895 & 0.837 & 0.623 & 0.644 & 0.516 \\
    TCC~\cite{shen2021you}   & 0.790 & 0.906 & 0.733 & 0.479 & 0.491 & 0.312 & 0.848 & 0.897 & 0.825 & 0.554 & 0.595 & 0.417 \\
    \midrule
    SimCLR~\cite{chen2020simple} & 0.658 & 0.753 & 0.601 & 0.386 & 0.358 & 0.239 & 0.796 & 0.865 & 0.753 & 0.339 & 0.332 & 0.189 \\
    MoCo~\cite{he2020momentum}  & 0.669 & 0.776 & 0.608 & 0.390 & 0.397 & 0.242 &   --    &    --   &  --     & 0.347 & 0.338 & 0.197 \\
    SimSiam~\cite{chen2021exploring} & 0.786 & 0.856 & 0.736 & 0.522 & 0.485 & 0.327 & 0.831 & 0.921 & 0.833 & 0.583 & 0.674 & 0.501 \\
    BYOL~\cite{grill2020bootstrap}  & 0.817 & 0.894 & 0.790 & 0.559 & 0.569 & 0.393 & 0.866 & 0.939 & 0.872 & 0.635 & 0.694 & 0.548 \\
    C-BYOL~\cite{lee2021compressive} & 0.823 & 0.901 & 0.804 & 0.564 & 0.575 & 0.408 & 0.859 & 0.943 & 0.885 & 0.629 & 0.687 & 0.539 \\
    MoCo-CLD~\cite{wang2021unsupervised} &    0.724 & 0.783 & 0.620 & 0.398 & 0.403 & 0.251 & 0.815 & 0.872 & 0.761 & 0.353 & 0.342 & 0.213 \\
    BYOL-CLD~\cite{wang2021unsupervised} & 0.826 & 0.902 & 0.808 & 0.563 & 0.578 & 0.408 & 0.874 & 0.948 & 0.880 & 0.648 & 0.703 & 0.555 \\
    NNCLR~\cite{dwibedi2021little} & 0.812 & 0.897 & 0.791 & 0.558 & 0.571 & 0.392 & 0.868 & 0.939 & 0.871 & 0.638 & 0.695 & 0.549 \\\midrule
    IDFD~\cite{tao2021clustering}  & 0.711 & 0.815 & 0.663 & 0.426 & 0.425 & 0.264 & \underline{0.898} & 0.954 & 0.901 & 0.546 & 0.591 & 0.413 \\
    PCL~\cite{li2020prototypical}   & 0.802 & 0.874 & 0.766 & 0.528 & 0.526 & 0.363 & 0.841 & 0.907 & 0.822 & 0.440 & 0.412 & 0.299 \\
    ProPos~\cite{huang2022learning} & \underline{0.886} & \underline{0.943} & \underline{0.884} & \underline{0.606} & \textbf{0.614} & \underline{0.451} & 0.896 & \underline{0.956} & \underline{0.906} & \underline{0.692} & \underline{0.745} & \underline{0.627} \\
    PIPCDR (Ours) & \textbf{0.897} & \textbf{0.948} & \textbf{0.894} & \textbf{0.623} & \underline{0.601} & \textbf{0.456} & \textbf{0.900} & \textbf{0.959} & \textbf{0.911} & \textbf{0.745} & \textbf{0.765} & \textbf{0.680} \\
    \bottomrule
    \end{tabular}%
  \label{tab:2}%
\end{table*}%

In this section, We perform a thorough evaluation of PIPCDR, comparing it to the state-of-the-art clustering approaches on multiple benchmark datasets. To conduct a systematic analysis, we categorize these approaches into five distinct types: 
\begin{itemize}
\item[i)] Methods that do not employ \ac{cl}: IIC~\cite{ji2019invariant},  DCCM~\cite{wu2019deep} and PICA~\cite{huang2020deep};
\item[ii)] Multi-stage methods that necessitate step-by-step pretraining or fine-tuning: SCAN~\cite{van2020scan}, and NMM~\cite{dang2021nearest};
\item[iii)] Methods that directly output cluster assignments: CC~\cite{li2021contrastive}, MiCE~\cite{tsai2021mice}, GCC~\cite{zhong2021graph}, TCL~\cite{li2022twin}, and TCC~\cite{shen2021you};
\item[iv)] Methods that focus on learning general representations: SimCLR~\cite{chen2020simple}, MoCo~\cite{he2020momentum}, SimSiam~\cite{chen2021exploring}, BYOL~\cite{grill2020bootstrap}, and C-BYOL~\cite{lee2021compressive};
\item[v)] Methods specifically designed to enhance \ac{rl} for clustering: IDFD~\cite{tao2021clustering}, PCL~\cite{li2020prototypical}, and ProPos~\cite{huang2022learning}.
\end{itemize}
To ensure fair comparisons, we meticulously adhere to the experimental settings employed in previous works \cite{tao2021clustering,tsai2021mice}. This rigorous approach guarantees the validity and reliability of our evaluation, enabling a meaningful assessment of the performance of PIPCDR in relation to the referenced studies. In all tables presenting comparison results, the best-performing method is prominently highlighted in bold, while the second-best method is distinguished with underlining. This visual emphasis aids in easily identifying and interpreting the superior performance of the top-ranked approach and the strong performance of the runner-up.

Table \ref{tab:2} presents the performance comparisons on four moderate-scale datasets. PIPCDR consistently outperforms other methods, showcasing the superiority of PIPCDR in \ac{dc} and its ability to capture semantic class information effectively. Given the relatively small size of the ImageNet-10 dataset, which contains only 13K images, our PIPCDR achieves competitive performance compared to ProPos~\cite{huang2022learning} and IDFD~\cite{tao2021clustering}. It is important to note that the limited size of this dataset may not allow for discriminative differences to emerge among contemporary state-of-the-art approaches. Over the ImageNet-Dogs dataset, which is a refined dataset consisting of various dog species from the ImageNet dataset, our PIPCDR demonstrates significant improvements of nearly $20\%$ over previous state-of-the-art methods except for the ProPos method. In the case of ProPos, PIPCDR outperforms it by delivering superior clustering results on the ImageNet-Dogs dataset. This dataset also poses challenges for contrastive-based methods due to the issue of severe class collision, where instances from the same class are pushed apart. However, IDFD shows some capability in addressing this problem to some extent through feature decorrelation and instance discrimination, and ProPos has made significant advancements by effectively addressing class-collision and uniform representation challenges. Furthermore, BYOL, C-BYOL, and SimSiam demonstrate notable improvements compared to SimCLR and MoCo. This highlights the significant potential of non-contrastive \ac{rl} for \ac{dc} without encountering the class collision issue. However, the proposed PIPCDR method combines the advantages of both contrastive and non-contrastive \ac{rl} approaches. It effectively tackles critical issues such as avoiding class collision, enhancing clustering stability through uniform representations, achieving non-overlapping clusters, and improving intra-cluster cohesion. Therefore, PIPCDR outperforms other methods, delivering superior performance.

\begin{table}[!ht]
    \centering
    \caption{Comparison of the clustering results in terms of NMI, ACC, and ARI of competing methods on the Tiny-ImageNet dataset.}
    \setlength{\tabcolsep}{17pt}
    \begin{tabular}{lccc}\toprule
     \multirow{2}[3]{*}{\textbf{Methods}} & \multicolumn{3}{c}{\textbf{Tiny-ImageNet}} \\\cmidrule{2-4}
         & NMI & ACC & ARI \\\midrule
    DCCM~\cite{wu2019deep} & 0.224 & 0.108 & 0.038 \\
    PICA~\cite{huang2020deep} & 0.277 & 0.098 & 0.040 \\
    CC~\cite{li2021contrastive}  & 0.340 &  0.140 & 0.071 \\
    GCC~\cite{zhong2021graph} & 0.347 & 0.138 & 0.075 \\
    MoCo~\cite{zhong2021graph} & 0.342 & 0.160 & 0.080 \\
    PCL~\cite{li2020prototypical} & 0.350 & 0.159 & 0.087 \\
    SimSiam~\cite{chen2021exploring} & 0.351 & 0.203 & 0.094 \\
    BYOL~\cite{grill2020bootstrap} & 0.365 & 0.199 & 0.100 \\
    ProPos~\cite{huang2022learning} & \underline{0.405} & \underline{0.256} & \underline{0.143} \\
    PIPCDR (Ours) & \textbf{0.479} & \textbf{0.331}   & \textbf{0.202}  \\\bottomrule
    \end{tabular}    
    \label{tab:3}
\end{table}

\begin{table}[!ht]
    \centering
    \setlength{\tabcolsep}{20pt}
    \caption{Comparison of the clustering results in terms of AMI of competing methods on the ImageNet-1k dataset.}
    \begin{tabular}{lc}\toprule
     \textbf{Method}    &  \textbf{AMI}\\\midrule
      DeepCluster~\cite{caron2018deep}   & 0.281 \\
      MoCo~\cite{zhong2021graph} & 0.285 \\
      PCL~\cite{li2020prototypical} & 0.410 \\
      ProPos~\cite{huang2022learning} & \underline{0.525} \\
      PIPCDR (Ours) & \textbf{0.542} \\\bottomrule
    \end{tabular} 
    \label{tab:4}
\end{table}

We evaluate the efficacy of PIPCDR on large-scale datasets that consist of a substantial number of classes. The evaluations are conducted on Tiny-ImageNet, which contains 200 classes, and ImageNet-1k, which consists of 1000 classes. We present the evaluation outcomes in Tables \ref{tab:3} and \ref{tab:4}, respectively. It should be noted that the tables only include methods that have reported results on the corresponding datasets. In the case of the ImageNet-1k dataset, we precisely adhered to the settings outlined in \cite{li2020prototypical}. We assess the performance using Adjusted Mutual Information (AMI), and the results of other algorithms reported in Table \ref{tab:4} are referenced from \cite{li2020prototypical}. Additionally, a ResNet-50 model is trained for 200 epochs, consistent with the methodology employed in \cite{li2020prototypical}. The outcomes demonstrate the robust versatility of PIPCDR on challenging datasets containing a substantial number of clusters.

According to the experimental results, we can conclude that PIPCDR consistently outperforms other methods in terms of clustering performance. It achieves improved within-cluster compactness, generates well-separated clusters, and demonstrates competitive performance on moderate-scale datasets. Furthermore, it achieves groundbreaking performance records on large-scale datasets, showcasing its effectiveness and superiority in the field of \ac{dc}. 

\subsection{Empirical analysis of PIP and CDR}
We present empirical analysis to justify how the proposed PIPCDR enhances \ac{rl} for \ac{dc} by considering the contributions of both PIP and CDR.
\subsubsection{Contribution of PIP}
To assess the validity of the assumption made by PIP that the sampled neighbors are indeed positive examples from the same semantic classes, we examine the behavior of PIP during training by monitoring any changes in the semantic classes of the input examples.
\begin{figure}
    \centering
    \includegraphics[width=0.7\linewidth]{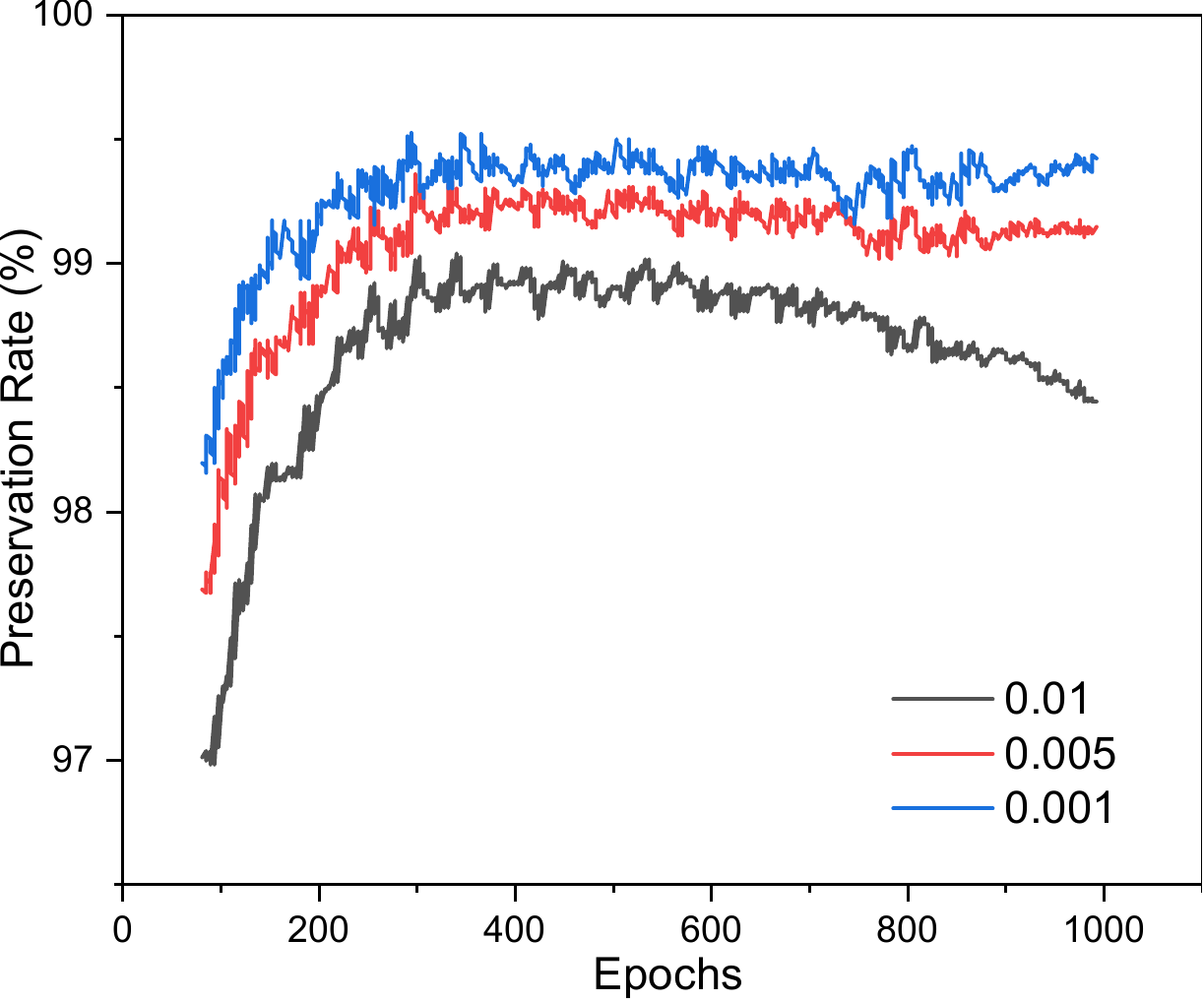}
    \caption{Comparison of preservation rate of sampled neighbors under different $\sigma$ during training.}
    \label{fig:5}
\end{figure}
To evaluate the preservation of original classes in the sampled neighbors, we conduct $k$-NN classification to predict the classes of both inputs and their sampled neighbors using a testing set after $100$ epochs. The preservation rate is determined by the proportion of sampled neighbors that maintain their original classes. We repeat the experiments $10$ times, varying the value of $\sigma$ for PIP over the CIFAR-10 dataset. In Fig. \ref{fig:5}, it can be observed that even in the early stages of training, the sampled neighbors exhibit a high preservation rate for $\sigma < 0.005$. However, the preservation rate decreases as expected for larger $\sigma$ values, such as $0.01$, as PIP may sample instances from other clusters. Based on the results, it can be concluded that the sampled neighbors are indeed positive examples, both quantitatively and qualitatively. This validation supports the effectiveness of PIP in enhancing within-cluster compactness.

\subsubsection{Contribution of CDR}
It is crucial to prevent the collapse of downstream clustering, where a large portion of samples are assigned to a few clusters due to the non-uniform representations generated by non-contrastive \ac{rl}. Therefore, it is imperative to strive for non-trivial solutions in \ac{dc}.

Our PIPCDR enhances \ac{rl} in the context of \ac{dc} by promoting cluster uniformity through CDR and leveraging the power of BYOL, a prominent \ac{ncl} method. In line with \cite{chen2021exploring}, we adopt the standard deviation (STD) of $\ell_2$-normalized representations as a standard of uniformity. When the $\ell_2$-normalized representations are evenly spread across a unit hypersphere, we anticipate the STD to approximate $1/\sqrt{d}$, where $d$ refers to the dimensionality of the feature representation $z$ and $z'$ represents its $\ell_2$-normalized version. To validate the significance of PIPCDR, we execute a series of experiments focusing on the uniformity of representations, the potential collapse of clustering, and the overall clustering performance.
\begin{figure}
    \centering
    \includegraphics[width=0.7\linewidth]{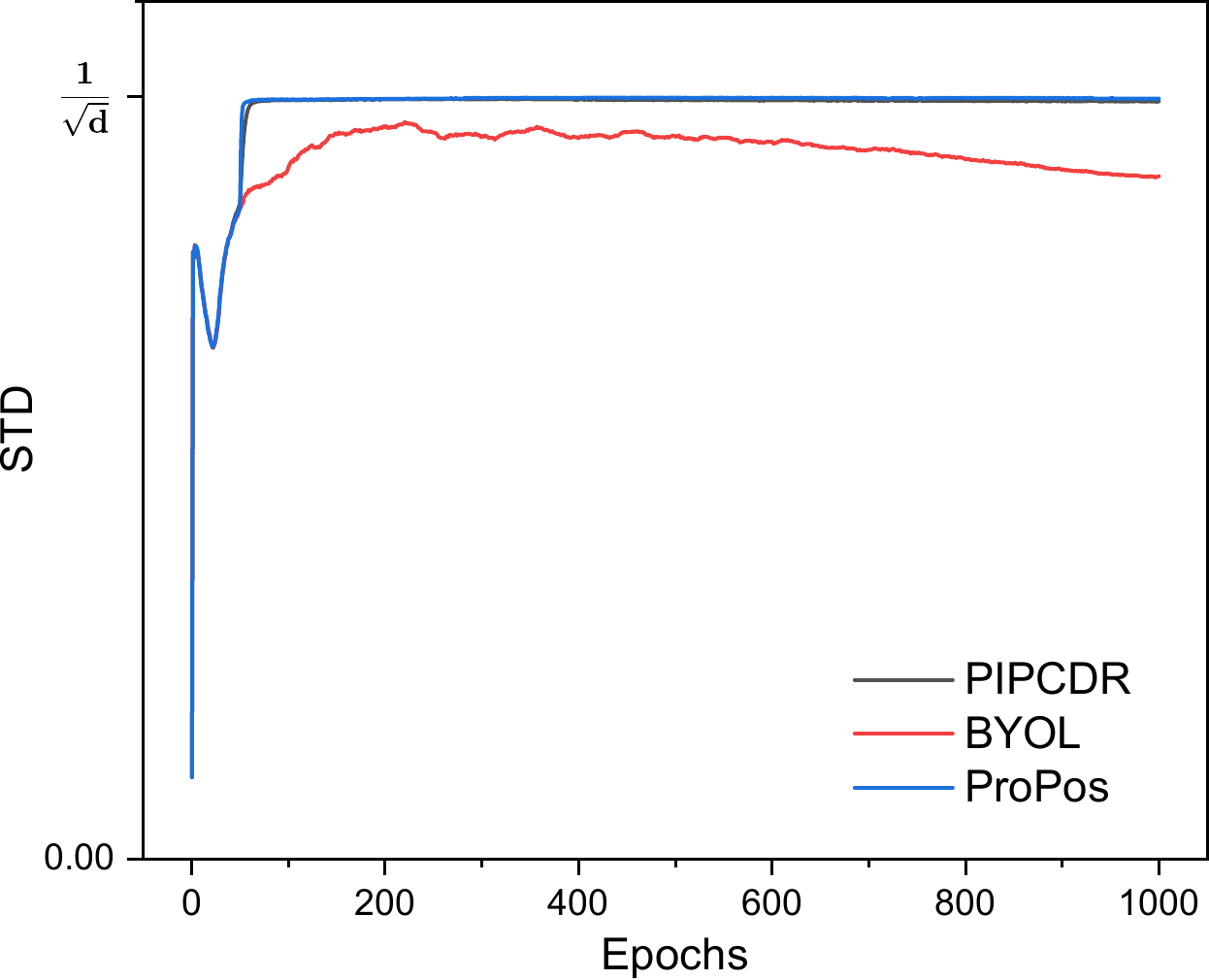}
    \caption{Comparison of standard deviation (STD) of $\ell_2$-normalized features to evaluate the uniformity.}
    \label{fig:1a}
\end{figure}
We begin by visualizing the uniformity of representations in Fig. \ref{fig:1a}. Upon examining the STD over the CIFAR-10 dataset during the training stage, we observe that PIPCDR and ProPos exhibit higher STDs, whereas BYOL displays instability with gradually decreasing STDs. It is worth noting that an STD nearer to $1/\sqrt{d}$ indicates a greater degree of uniformity in the representations. Notably, ProPos and PIPCDR approaches yield more uniform representations compared to BYOL. Furthermore, the uniformity of PIPCDR and ProPos remains relatively stable throughout the training process.
\begin{figure}
    \centering
    \includegraphics[width=0.7\linewidth]{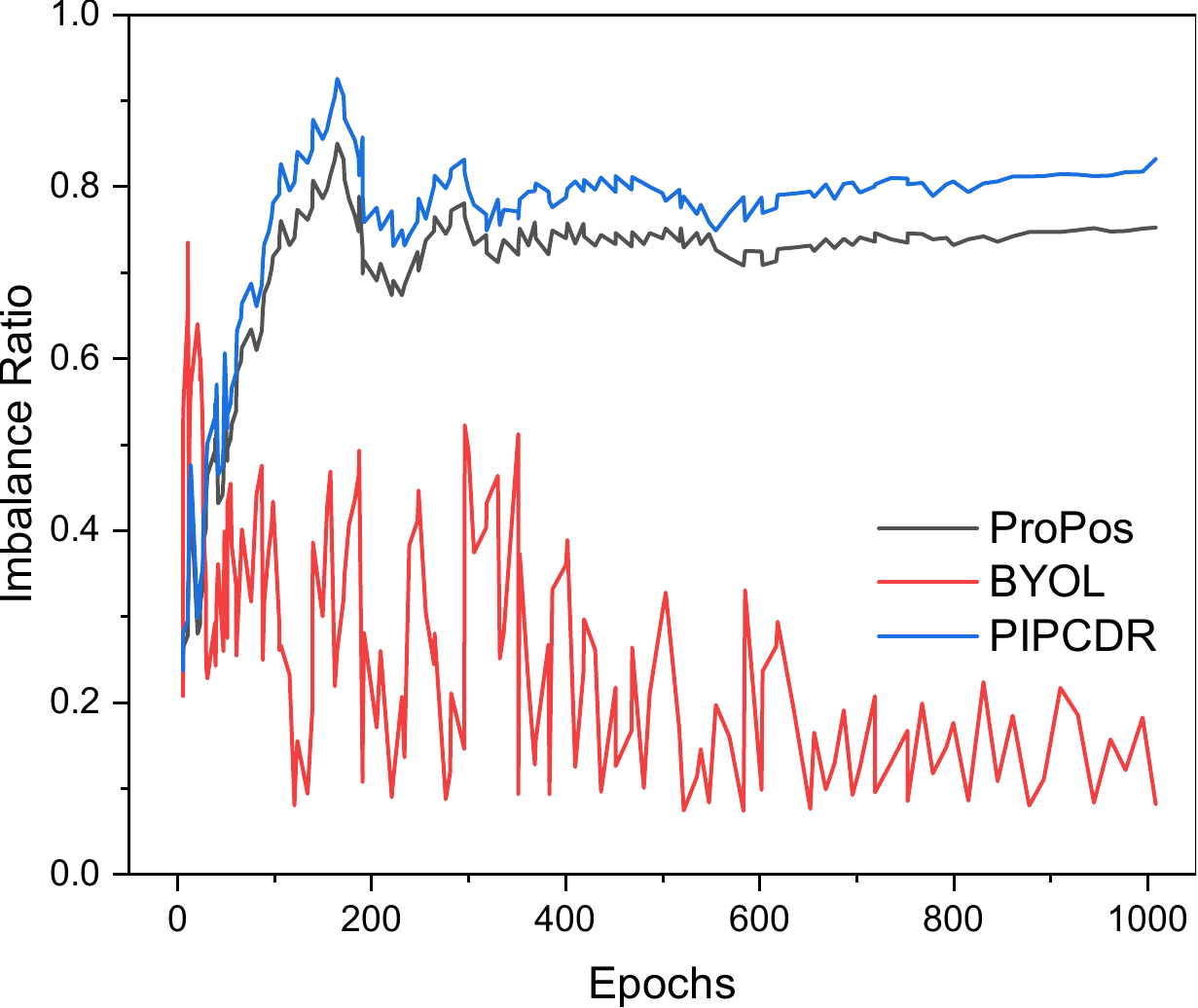}
    \caption{Comparison of cluster imbalance ratio to evaluate the cluster balance.}
    \label{fig:2}
\end{figure}
\begin{figure}
    \centering
    \includegraphics[width=0.7\linewidth]{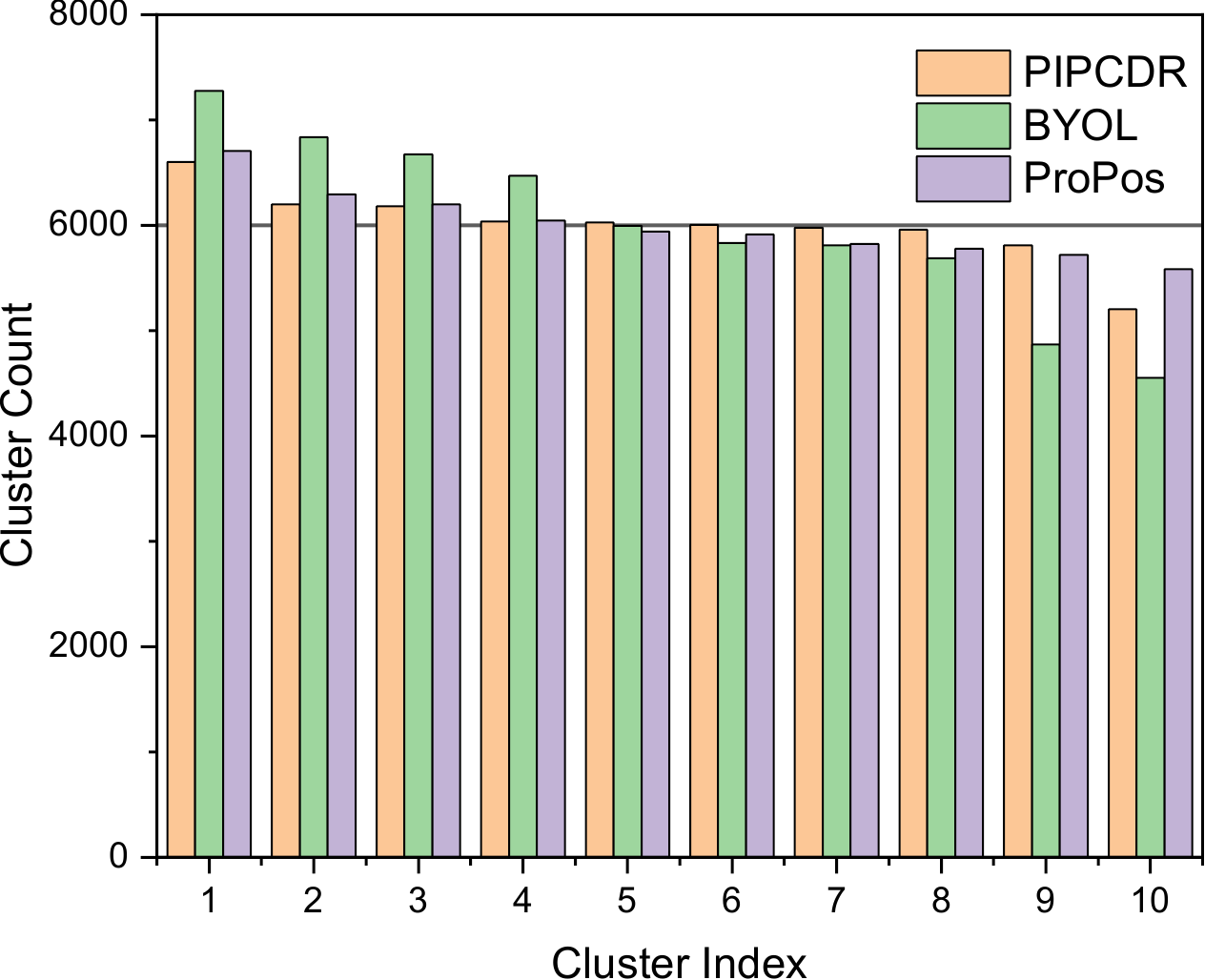}
    \caption{Comparison of cluster statistics, specifically the number of samples in each cluster, sorted in ascending order, obtained from the model at the 1000th epoch.}
    \label{fig:4}
\end{figure}

Given that PIPCDR and ProPos exhibit comparable levels of representation uniformity, we proceed to visualize the cluster imbalance as a means to assess the potential for clustering collapse. To provide a more precise analysis of our results, we employ a metric known as the cluster imbalance ratio. This metric serves as a quantitative measure of the disparity between the cluster containing the fewest samples and the cluster with the most samples within the dataset. Specifically, we calculate the ratio as $\min\left(\{N_k\}_{k=1}^K\right)/\max\left(\{N_k\}_{k=1}^K\right)$, where $N_k$ represents the total number of samples belonging to the $k$-th cluster. A higher value signifies a more equitable distribution of samples among clusters. Furthermore, we present the cluster statistics, precisely the sorted count of samples per cluster, for the model at the 1000th epoch. We depict the outcomes pertaining to cluster imbalance throughout training and the cluster accuracy at the end of each epoch in Fig. \ref{fig:2} and \ref{fig:4}, respectively. Fig. \ref{fig:2} demonstrates that the $k$-means clustering procedure employed by PIPCDR yields more balanced clusters, as evidenced by a higher cluster imbalance ratio compared to BYOL and ProPos. In contrast, the clusters obtained from BYOL exhibit a notable imbalance, which aligns with the decreasing STDs. Furthermore, Fig. \ref{fig:4} illustrates that the cluster statistics for PIPCDR demonstrate a nearly equal distribution across different clusters, in contrast to the other methods. The presence of more balanced clusters except for one class confirms that PIPCDR effectively mitigates the collapse of $k$-means clustering, surpassing both ProPos and BYOL.

\subsubsection{Combined contribution of PIP and CDR}
\begin{figure}
    \centering
    \includegraphics[width=0.7\linewidth]{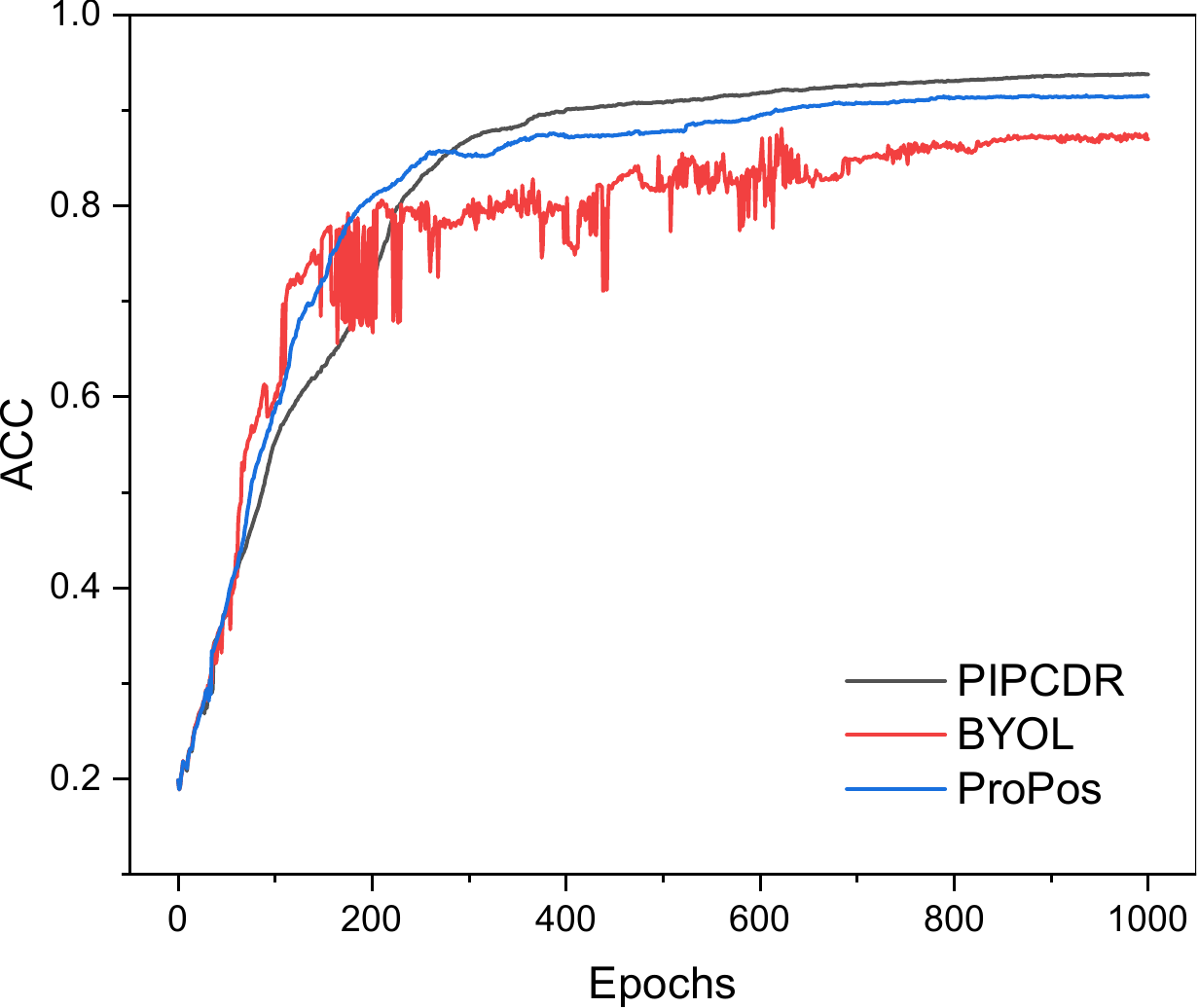}
    \caption{Comparison of accuracy to evaluate the clustering performance.}
    \label{fig:3}
\end{figure}

Finally, in Fig. \ref{fig:3}, we present the comparison of clustering performance among PIPCDR, ProPos, and BYOL, as measured by the accuracy (ACC) between the ground-truth labels and obtained clustering results. PIPCDR consistently achieves higher and more stable ACC values compared to ProPos and BYOL. Based on the analysis conducted, it can be inferred that although BYOL effectively addresses the class collision problem, it faces challenges with the collapse of $k$-means clustering caused by non-uniform representations. ProPos mitigates this issue through the incorporation of PSL, resulting in improved ACC compared to BYOL. In contrast, PIPCDR combines PIP and CDR to achieve greater uniformity in representations, well-organized sample clusters, and improved intra-cluster compactness compared to ProPos.

\subsection{Application on Long-tail Dataset}
\begin{table}[!ht]
    \centering
    \setlength{\tabcolsep}{13pt}
    \caption{Clustering outcomes in terms of NMI, ACC, and ARI of PIPCDR on the long-tail dataset. We show the best and second-best results in bold and underlined, respectively}
    \begin{tabular}{lccc}\toprule
       \multirow{2}[3]{*}{\textbf{Method}}  &  \multicolumn{3}{c}{\textbf{CIFAR-10-LT}} \\\cmidrule{2-4}
       & NMI & ACC & ARI \\\midrule
       MoCo~\cite{he2020momentum} & 0.467 & 0.334 & 0.277 \\
       BYOL~\cite{grill2020bootstrap} & 0.516 & 0.413 & 0.308 \\
       ProPos~\cite{huang2022learning} & 0.553 & 0.439 & \underline{0.363}\\\midrule
       PIP & 0.548 & 0.438 & 0.322 \\
       CDR & \underline{0.556} & \underline{0.442} & 0.353 \\
       PIPCDR & \textbf{0.568} & \textbf{0.456} & \textbf{0.373} \\\bottomrule\bottomrule
       \multirow{2}[3]{*}{\textbf{Method}}  &  \multicolumn{3}{c}{\textbf{CIFAR-20-LT}} \\\cmidrule{2-4}
       & NMI & ACC & ARI \\\midrule
       MoCo~\cite{he2020momentum} & 0.312 & 0.282 & 0.161 \\
       BYOL~\cite{grill2020bootstrap} & 0.419 & 0.346 & 0.223 \\
       ProPos~\cite{huang2022learning} & 0.446 & \underline{0.390} & 0.273\\\midrule
       PIP & 0.447 & 0.358 & 0.258 \\
       CDR & \underline{0.453} & 0.386 & \underline{0.283} \\
       PIPCDR & \textbf{0.463} & \textbf{0.405} & \textbf{0.294} \\\bottomrule
    \end{tabular}
    \label{tab:new_1}
\end{table}
To evaluate the robustness of our proposed PIPCDR algorithm, we conducted additional experiments on long-tailed datasets, as shown in Table \ref{tab:new_1}. In these experiments, we leveraged the long-tailed versions of CIFAR-20 and CIFAR-10, denoted as CIFAR-20-LT and CIFAR-10-LT, respectively. These datasets were constructed following the methodology outlined in \cite{tang2020long}, which builds upon \cite{zhou2020bbn} and \cite{cao2019learning}. Specifically, they were created with an intentional data imbalance ratio, where $\min({N_k}^K_{k=1})/ \max({N_k}^K_{k=1}) = 0.1$. Consequently, the majority of instances in the long-tailed datasets reside in the minority classes (head), while only a few samples are allocated to remaining classes (tail). Notably, MoCo v2 encounters challenges in handling this scenario due to class collision issues. As a consequence, it pushes the head samples away and mixes the tail samples together. In contrast, both BYOL ProPos and our PIPCDR algorithm excel in this context since they do not rely on negative examples, outperforming MoCo v2 significantly.

Building on the foundation of PIP and CDR, our PIPCDR algorithm further enhances clustering performance, as demonstrated in Table \ref{tab:new_1}. We showcase results in terms of NMI, ACC, and ARI. The superior performance of PIPCDR on long-tail datasets underscores its adaptability and effectiveness in challenging real-world scenarios. This evaluation highlights the potential of PIPCDR as a robust and versatile deep clustering algorithm, particularly in scenarios characterized by data imbalance and class collision challenges.

\subsection{Utilization of Memory Queue}
\begin{table}[!ht]
    \centering
    \setlength{\tabcolsep}{18pt}
    \caption{Clustering outcome in terms of AMI on the ImageNet-1k and Tiny-ImageNet datasets for varying memory queue size.}
    \begin{tabular}{ccc}\toprule
    \multirow{2}[3]{*}{\textbf{Memory-size}} &  \multicolumn{2}{c}{\textbf{AMI}} \\\cmidrule{2-3}
      & ImageNet-1K & Tiny-ImageNet  \\\midrule
      0   &  0.542 & 0.468\\
      512 & 0.568 & 0.476\\
      1024 & \underline{0.571} & \textbf{0.498} \\
      2048 & \textbf{0.583} & \textbf{0.504}\\
      4096 & 0.551 & 0.472\\\bottomrule 
    \end{tabular} 
    \label{tab:new_2}
\end{table}

To accurately represent the cluster of a single class, it is essential to have a sufficient number of samples within the mini-batch. However, this requirement can significantly increase GPU memory demands, particularly when dealing with datasets that feature a large number of clusters. Notably, even with a modest mini-batch size of 256, our PIPCDR algorithm demonstrates strong generalization capabilities on datasets like Tiny-ImageNet, which consists of 200 classes (equating to approximately 1 sample per class in a mini-batch), and ImageNet-1k, with its 1,000 classes (about 0.25 sample per class).

To enhance the precision of CDR loss computation with a small mini-batch size, we propose the use of a memory queue for cluster estimation. In the conventional CDR approach, cluster estimation for calculating CDR loss relies solely on representations within the mini-batch $\mathcal{B}$, as outlined in Eqn (\ref{eqn_7}). We adopt a memory queue $\mathcal{Q}$, a concept resembling of prior work in \cite{he2020momentum}, where representations from the momentum-updated encoder are stored. This approach extends cluster estimation to cover samples drawn from both the current mini-batch ($\mathcal{B}$) and the memory queue ($\mathcal{Q}$), creating a unified representation set ($\mathcal{B} \cup \mathcal{Q}$). 

To illustrate the proficiency of our memory queue approach, we conduct rigorous assessments on challenging datasets, including ImageNet-1k and Tiny-ImageNet, featuring 1000 and 200 classes, respectively. The outcomes, which are detailed in Table \ref{tab:new_2}, demonstrate the remarkable potential of our model, trained under the same settings as our primary experiments. As depicted in Table \ref{tab:new_2}, we observe a clear performance improvement when we incorporate more samples for cluster computation. This underscores the significance of the memory queue in enriching the representation learning process. However, a critical observation emerges: excessively enlarging the memory queue, as indicated by the case with 4,096 entries in Table \ref{tab:new_2}, can paradoxically lead to a decline in performance. This phenomenon arises due to the longer encoder updates associated with the larger memory queue, which can cause the representations enqueued during the early iterations to deviate significantly from the true representations \cite{he2020momentum}. Moreover, it's essential to acknowledge the trade-offs that come into play. The utilization of a memory queue introduces additional computational costs, and the determination of an appropriate hyperparameter—the size of the memory queue—becomes a critical consideration.

In summary, our proposed PIPCDR algorithm exhibits exceptional performance, particularly when dealing with small mini-batch sizes relative to the number of classes. Its effectiveness is further underscored by the judicious utilization of a memory queue, which not only enhances performance but also demands a thoughtful approach to hyperparameter selection and management. This comprehensive analysis reaffirms the robustness and versatility of our approach in the context of representation learning.

\subsection{Analysis of Effect of Hyperparameters}
\begin{figure}
    \centering
    \includegraphics[width=0.7\linewidth]{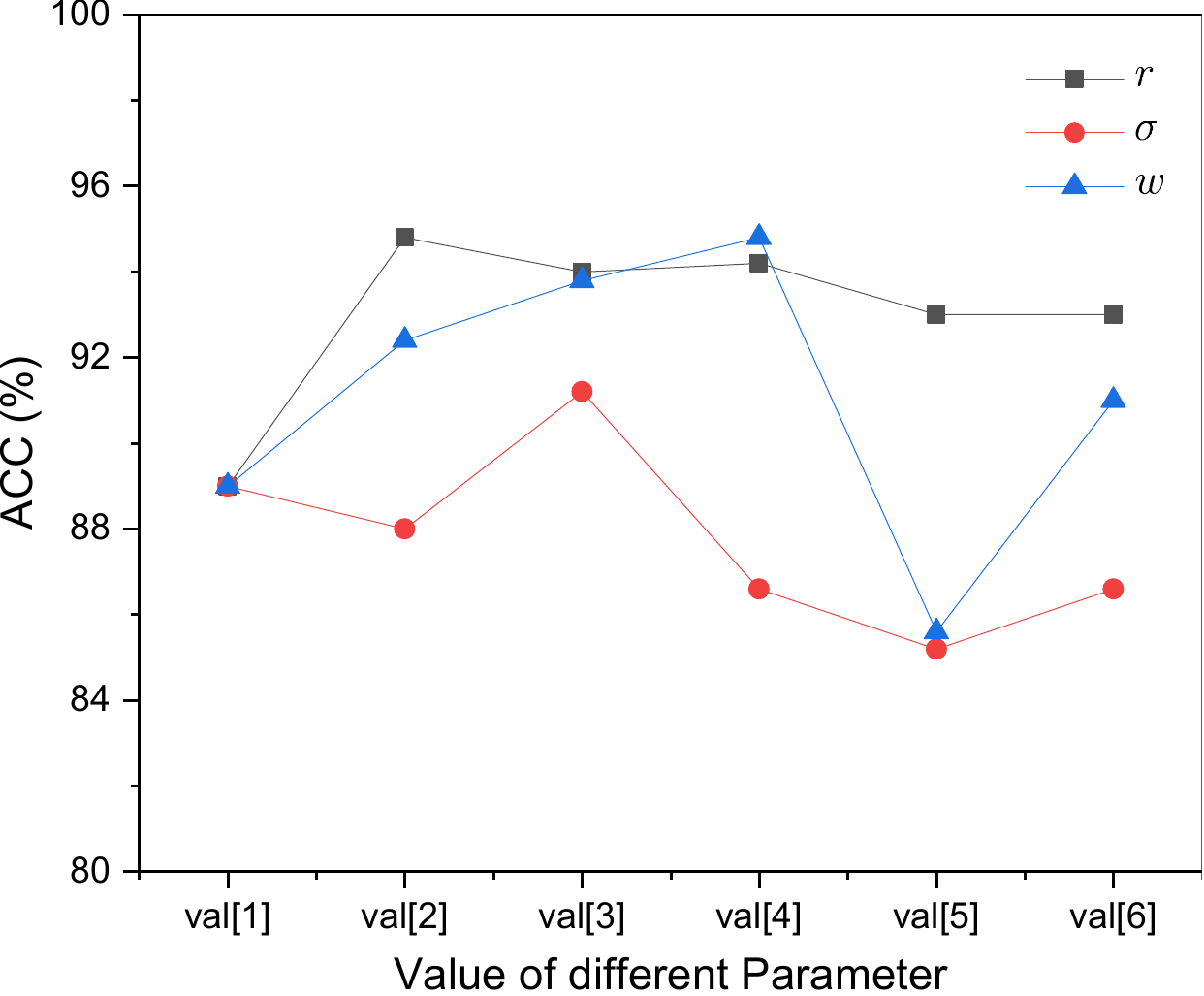}
    \caption{Effect of different parameters on the performance of PIPCDR. Here, the values of val[$\cdot$] for $r$, $\sigma$, and $w$ are as follows: $\{0,1,2,4,8,16\}$, $\{0, 0.0001, 0.001, 0.005, 0.01, 0.05\}$, and $\{0, 0.01, 0.05, 0.1, 0.5, 1\}$, respectively.}
    \label{fig:8}
\end{figure}
In order to examine the impact of various hyperparameters, we conducted comprehensive experiments on the CIFAR-10 datasets. In self-supervised learning, we employed BYOL and ProPos as the baseline methods for the projection dimension, backbone, and data augmentation.
\subsubsection{$k$-means Clustering Frequency}
Our investigation focuses on the impact of varying $r$ values on the clustering performance. In the context of PIPCDR, we perform $k$-means clustering at every $r$ epoch. The results presented in Fig. \ref{fig:8} indicate the robustness of PIPCDR to large $r$ values and the cluster pseudo-labels. These findings imply that conducting clustering for every epoch is not essential, thereby leading to a substantial reduction in computational costs. After conducting a thorough analysis, it can be inferred that setting the value of $r$ within the range of $[1, 16]$ yields commendable performance for the datasets under consideration.
\subsubsection{Impact of $\sigma$}
We can control the extent of positive sampling towards neighboring instances by adjusting the hyperparameter $\sigma$ in PIP. Analyzing the range of $\sigma \sim [0, 10^{-3}]$ as shown in Fig. \ref{fig:8}, we notice that incorporating positive sampling into BYOL results in a minor performance decrease on the CIFAR-10 dataset. However, it significantly improves the stability of the clustering performance, as indicated by the reduced standard deviation. This improvement can be attributed to considering the neighbors of each sample as positive examples. These findings suggest that positive sampling contributes to enhanced performance stability. Nevertheless, when $\sigma$ is set to a large value, the performance becomes unstable and experiences a significant decline. This outcome is expected because, with large $\sigma$ values, samples from distinct clusters might be sampled and incorrectly treated as positive instances. Thus, it is advisable to opt for a small value of $\sigma$, preferably within the range of $(0, 10^{-3}]$, to ensure adequate positive sampling while maintaining stability in performance.
\subsubsection{Impact of $w$}
The hyperparameter $w$ plays a crucial role in determining the impact of CDR on the overall performance. The hyperparameter $w$ controls the balance between PIP and CDR during training. In Fig. \ref{fig:8}, we present the results of varying $w$ and its corresponding effect on the clustering performance over the CIFAR-10 dataset. Notably, higher values of the hyperparameter $w$ tend to introduce instability, resulting in suboptimal outcomes. In order to achieve robust and reliable performance, it is advisable to set the value of $w$ within the range of $[0.01, 0.1]$. This range has consistently exhibited superior performance on a given dataset, providing a balance between the utilization of CDR and overall stability. In conclusion, the hyperparameter $w$ holds significance in the PIPCDR framework, allowing practitioners to control the importance of CDR and strike a balance between stability and performance.

\subsection{Ablation Study}
In this study, we conduct comprehensive ablation studies encompassing both qualitative and quantitative assesssments. We aim to delve deeper into the underlying factors contributing to the robustness of PIPCDR in \ac{dc}. Through this analysis, we aim to shed light on the precise mechanisms and insights that underscore the efficacy of PIPCDR.
\subsubsection{Ablation study of PIP and CDR}
\begin{table}[!ht]
    \centering
    \setlength{\tabcolsep}{13pt}
    \caption{Ablation studies in terms of NMI, ACC, and ARI of PIP and CDR for PIPCDR. We show the best and second-best results in bold and underlined, respectively}
    \begin{tabular}{lccc}\toprule
       \multirow{2}[3]{*}{\textbf{Method}}  &  \multicolumn{3}{c}{\textbf{CIFAR-10}} \\\cmidrule{2-4}
       & NMI & ACC & ARI \\\midrule
       BYOL & 0.817 & 0.894 & 0.790 \\\midrule
       PIPCDR & \textbf{0.897} & \textbf{0.948} & \textbf{0.894} \\
       PIP & 0.816 & 0.897 & 0.803 \\
       CDR & 0.858 & 0.925 & 0.847 \\
       PIPCDR w/o uniformity & 0.863 & 0.931 & 0.856 \\
       PIPCDR w/o alignment & \underline{0.885} & \underline{0.936} & \underline{0.877} \\\bottomrule
    \end{tabular}
    \label{tab:5}
\end{table}
 We propose four distinct variants: PIP, CDR, PIPCDR w/o alignment, and PIPCDR w/o uniformity to investigate further the impact of individual components in the PIPCDR framework. We design each variant by selectively removing specific elements from the original framework, namely PIP, CDR, PIPCDR w/o uniformity, and PIPCDR w/o alignment, respectively. Through these variants, we aim to gain a comprehensive understanding of the contribution and significance of each component in the overall performance of PIPCDR. We depict the outcomes of these variants alongside PIPCDR on the CIFAR-10 dataset in Table. \ref{tab:5}.

 As shown in Table. \ref{tab:5}, CDR significantly enhances the baseline results, surpassing the marginal improvements achieved by PIP. This observation highlights the crucial role of CDR in enhancing the performance in clustering tasks. Nevertheless, PIP also has a crucial impact in two aspects. Firstly, incorporating PIP alone not only stabilizes the performance but also demonstrates further improvements compared to the baseline BYOL. Secondly, when combined with CDR in PIPCDR, PIP enhances the clustering process by facilitating the selection of positive neighbors. This is particularly beneficial because CDR ensures well-separated clusters, enabling PIP to sample positive neighbors belonging to the same semantic classes. Moreover, CDR considers both inter-cluster distance and within-cluster compactness, making it a comprehensive approach. As a result, the combination of PIP and CDR yields optimal clustering results, as CDR aligns instances within the same cluster to enhance within-cluster compactness while maximizing the inter-cluster distance, and PIP further enhances intra-cluster cohesion.

 To delve deeper into the impact of CDR, we divide it into two components: alignment and uniformity. We refer to these components as CDR w/o uniformity and CDR w/o alignment, respectively, as shown in Table. \ref{tab:5}. The results clearly indicate that the alignment term provides only marginal performance improvement, while the uniformity term contributes significantly to the overall gain. It is worth noting that when considering only the alignment term, our approach involves calculating the loss after the predictor network, as opposed to the feature extractor. This premeditated choice is made to prevent the risk of representation collapse. This observation suggests that uniformity plays a more crucial role than alignment in scattering the clusters and promoting uniform representations, which effectively addresses the issues discussed previously. Nevertheless, the alignment term remains essential for stabilizing the training process.

\subsubsection{Impact of $K$}
\begin{figure}
    \centering
    \includegraphics[width=0.7\linewidth]{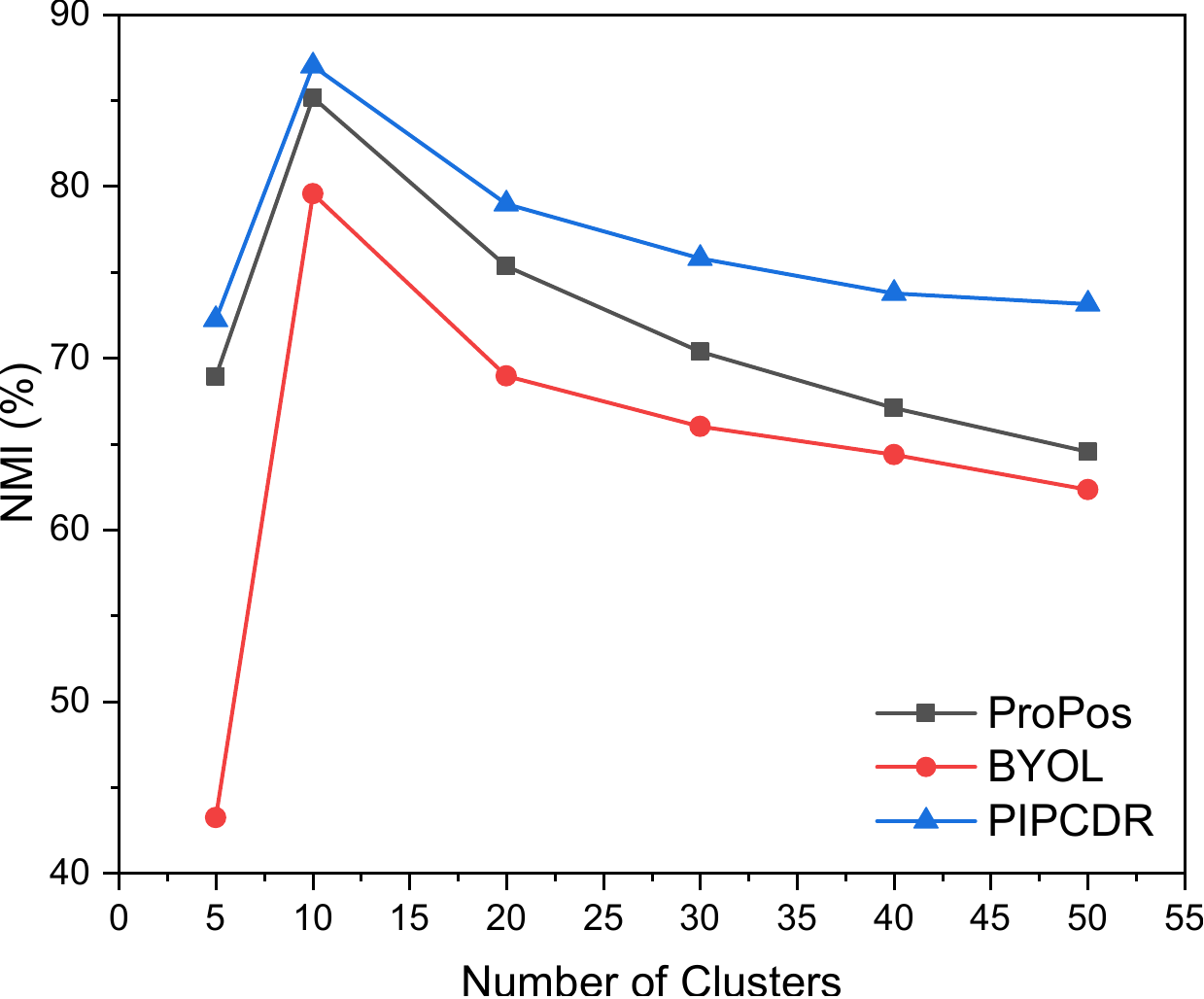}
    \caption{Comparison of NMI under a different predefined number of clusters $K$.}
    \label{fig:6}
\end{figure}
The abovementioned experiments utilize the total number of clusters, denoted by $K$, as a predefined parameter determined based on the ground truth. However, this approach may only be applicable in practical situations where the number of clusters is known. To address this,  we apply BYOL, PIPCDR, and ProPos methods on the CIFAR-10 dataset, considering varying numbers of clusters denoted as $K \in \{5, 10, 20, 30, 40, 50\}$. We present the NMI scores in Fig. \ref{fig:6}. It is important to note that the predetermined number of clusters ($K$) used in PIPCDR and ProPos during training aligns with the $K$ used in the subsequent $k$-means operations. Additionally, to conduct a comprehensive analysis of the impact of $K$, the outcomes of standard BYOL are also incorporated within the $k$-means procedure.

In scenarios where the number of clusters ($K$) is smaller than the actual number of classes (under-clustering), the clustering performance of all methods is noticeably affected. However, compared to BYOL, both PIPCDR and ProPos exhibit superior performance in these cases. This can be attributed to the enhanced uniformity in representation provided by both methods, as discussed in Section 4.3. In cases of over-clustering (when $K$ exceeds the number of ground-truth classes), the performance of all algorithms tends to decline. However, even in such scenarios, our PIPCDR algorithm consistently outperforms both BYOL and ProPos. The critical factor behind this improvement lies in the fact that PIPCDR not only ensures alignment within the same semantic clusters but also enhances uniformity in the representations. Over-clustering attempts to disrupt the cluster structures without adequate uniformity and push away semantically similar examples, ultimately compromising the clustering performance. This is the primary reason for the significant performance drop observed in BYOL. Furthermore, PIPCDR surpasses ProPos due to its ability to achieve a better cluster imbalance ratio during training, as explained in Section 4.3.

\section{Conclusion}
In this work, we introduce PIPCDR, an innovative \ac{dc} approach that combines the advantages of both contrastive- and non-contrastive-based approaches. It effectively addresses the issue of class collision, ensuring clustering stability, distinct clusters, and compactness within each cluster. In this method, we improve within-cluster compactness and achieve enhanced clustering outcomes by incorporating the \ac{pip} loss. Furthermore, the innovative \ac{cdr} aligns clusters by minimizing intra-cluster distance and maximizing between-cluster scattering, promoting compactness and separation. Extensive experiments on benchmark datasets demonstrate PIPCDR's superior performance, outperforming state-of-the-art methods, especially on large-scale datasets. These results affirm the efficacy and competitive nature of the proposed approach. The successful implementation of PIPCDR in clustering demonstrates its potential for future applications in image classification, object detection, and image segmentation. This method's robustness and versatility open doors to enhanced performance and novel advancements in diverse fields.

\section*{Acknowledgments}
This work was supported by the National Research Foundation of Korea (NRF) grant funded by the Korea government (MSIT) (No.2021R1C1C1012590).





\ifCLASSOPTIONcaptionsoff
  \newpage
\fi

\bibliographystyle{IEEEtran}
\bibliography{bibtex/bib/IEEEexample}




\end{document}


\newacro{dcnn}[DCNN]{deep convolutional neural network}
\newacro{nas}[NAS]{neural architecture search}
\newacro{dc}[DC]{deep clustering}
\newacro{cl}[CL]{contrastive learning}
\newacro{ncl}[NCL]{non-contrastive learning}
\newacro{rl}[RL]{representation learning}
\newacro{pip}[PIP]{positive instances proximity}
\newacro{cdr}[CDR]{cluster dispersion regularizer}
\newacro{mm}[MM]{majorize-minimization}
\title{\textit{Supplementary Document for -} ``Enhancing Clustering Representations with Positive Proximity and Cluster Dispersion Learning"\\
\vspace{5 mm}
\large (For online publication only)}

\author{Abhishek~Kumar,
        and~Dong-Gyu~Lee\IEEEauthorrefmark{2}
\thanks{A. Kumar, and D.G. Lee are with the Department of Artificial Intelligence, Kyungpook National University, Daegu, Republic of Korea - 41566}
\thanks{\IEEEauthorrefmark{2} Corresponding author (e-mail: dglee@knu.ac.kr).}
\thanks{Manuscript received XXXXX, 2023; revised August XXXXX, 2023.}}

\maketitle

\thispagestyle{empty}
 \renewcommand{\thetable}{\Roman{table}}
\renewcommand{\thefigure}{\Roman{figure}}

\section{Relation to Other Baselines}
In this section, we examine the differences between our methodology and the comparable baseline approaches.
\subsection{ProPos}
Both PIPCDR and ProPos~\cite{huang2022learning} integrate a uniformity term into the non-contrastive loss function to enhance the clustering performance. As outlined in Section 4.3, PIPCDR demonstrates superior performance compared to ProPos by effectively optimizing the cluster imbalance ratio during the training process. This enhancement directly results from the following key distinction between ProPos and PIPCDR.
\begin{itemize}[labelsep=-4.2cm, listparindent = 0ex, leftmargin = 0cm]
    \item[\textbf{Positive sampling process: }] \hspace{4.05cm} In the positive sampling process, ProPos generates a positive sample for a given instance by utilizing the following equation.
    \begin{equation}
        v_i = f(x_i^a) +\sigma \epsilon,~\mbox{where}~~ \epsilon \sim \mathcal{N}(0,I),
    \end{equation}
    where $\mathcal{N}$ represents Gaussian distribution. While the sampling strategy in ProPos contributes to performance improvement compared to vanilla BYOL, it does not take into account the actual nearest neighbors of the instances. In contrast, PIPCDR leverages the actual nearest neighbors in the positive sampling process, as demonstrated in Eq. (\ref{eqn_4}), ensuring that the samples form truly positive pairs and possess similar semantic characteristics. This approach significantly enhances within-cluster compactness, leading to improved clustering performance.
\end{itemize}
\begin{itemize}[labelsep=-4.32cm, listparindent = 0ex, leftmargin = 0cm]
    \item[\textbf{Class-level contrastive loss: }]\hspace{4.2cm} While ProPos applies the class-level contrastive loss to the representation of cluster centers, our approach directly applies the contrastive loss to the representation itself. This distinction is significant because implementing the contrastive loss solely on the representation of cluster centers, as done in \cite{huang2022learning}, fails to promote within-cluster compactness, even with the inclusion of an alignment term in the loss. However, within-cluster compactness plays a crucial role in facilitating effective \ac{rl} and enhancing performance in downstream tasks. In contrast, PIPCDR takes a different approach by implementing the contrastive loss on the mini-batch representations, utilizing the pseudo-labels generated from the $k$-means algorithm. This strategy enables PIPCDR to promote within-cluster compactness and effectively achieve highly discriminative cluster representations.
\end{itemize}
\begin{table}[!ht]
    \centering
    \setlength{\tabcolsep}{13pt}
    \caption{Comparison among PIPCDR, ProPos, BYOL, CC, and PCL. Here, PSA and PSL is part of the ProPos.}
    \begin{tabular}{lccc}\toprule
       \multirow{2}[3]{*}{\textbf{Method}}  &  \multicolumn{3}{c}{\textbf{CIFAR-10}} \\\cmidrule{2-4}
       & NMI & ACC & ARI \\\midrule
       CC~\cite{li2021contrastive} & 0.705 & 0.790 & 0.637 \\
       PCL~\cite{li2020prototypical} & 0.802 & 0.874 & 0.766 \\
       BYOL~\cite{grill2020bootstrap} & 0.817 & 0.894 & 0.790 \\
       ProPos~\cite{huang2022learning} & 0.886 & 0.943 & 0.884 \\
       PIPCDR & \textbf{0.897} & \textbf{0.948} & \textbf{0.894} \\\midrule
       CC+CDR & 0.754 & 0.852 & 0.701 \\
       BYOL+CC & 0.766 & 0.863 & 0.738 \\
       BYOL+PCL & 0.744 & 0.853 & 0.714 \\
       PIP+PSL & 0.889 & 0.944 & 0.887 \\
       PSA+CDR & 0.893 & 0.945 & 0.891 \\\midrule
       PSA & 0.794 & 0.879 & 0.764 \\
       PIP & 0.816 & 0.897 & 0.803 \\\midrule
       PSL & 0.834 & 0.903 & 0.811 \\
       CDR & 0.858 & 0.925 & 0.847 \\\midrule
       ProPos w/o uniformity & 0.796 & 0.878 & 0.765 \\
       PIPCDR w/o uniformity & 0.863 & 0.931 & 0.856 \\\midrule
       ProPos w/o alignment & 0.853 & 0.921 & 0.844 \\
       PIPCDR w/o alignment & 0.885 & 0.936 & 0.877 \\\bottomrule
    \end{tabular}
    \label{tab:6}
\end{table}

Furthermore, our experimental analysis includes a detailed comparison between the different components of ProPos and PIPCDR. The results, presented in Table \ref{tab:6}, provide a comprehensive evaluation of performance. Interestingly, the findings clearly indicate that both ProPos as a whole and its individual components yield lower performance compared to PIPCDR and its corresponding counterparts. Conversely, when evaluated under identical conditions, PIP+PSL and PSA+CDR demonstrate remarkable enhancements over ProPos, thus highlighting the superior performance of PIPCDR and its components in achieving significant improvements.

\subsection{Instance-reweighted Contrastive Loss}
In this section, we present a novel perspective to comprehend the proposed cluster-wise CDR based on the instance-reweighted contrastive loss~\cite{mitrovic2020representation}. Here, our primary focus centers on the analysis of the alignment term of CDR, which solely incorporates instances selected from the same class, resembling the supervised \ac{cl}~\cite{khosla2020supervised}. Hence, the CDR’s alignment term represents a comprehensive variant of instance-reweighted contrastive loss, incorporating the pseudo-labels. Likewise, it can be observed that the CDR’s uniformity term aims to enhance the inter-cluster gaps among samples originating from distinct classes.

Therefore, the proposed CDR can be comprehended as an extension of instance-reweighted contrastive loss, incorporating cluster pseudo-labels in its formulation.
\subsection{Contrastive Clustering Loss}
While both CDR and CC~\cite{li2021contrastive} utilize cluster-level contrastive loss for training, they exhibit distinct differences.
\begin{itemize}[labelsep=-4.52cm, listparindent = 0ex, leftmargin = 0cm]
    \item[\textbf{Cluster-level contrastive loss:}]\hspace{4.5cm}  In CC~\cite{li2021contrastive}, the cluster-level contrastive loss utilizes the cluster probabilities, whereas our approach applies it to the representation of individual instances. Implementing contrastive loss on the cluster probability in \cite{li2021contrastive} may result in the loss of interpretive insights in the acquired representations, which could be more conducive to the effective learning of representations. On the other hand, CDR takes a different approach by applying contrastive loss to the representations of individual instances inside a mini-batch, utilizing pseudo-labels derived from $k$-means algorithm. This enables CDR to capture the interpretive insights present in the latent space and generate representations that are further differentiating and well-suited for the process of clustering.
\end{itemize}
\begin{itemize}[labelsep=-2.98cm, listparindent = 0ex, leftmargin = 0cm]
    \item[\textbf{Cluster uniformity:}]\hspace{2.9cm} While the cluster-level contrastive loss in CC lacks the capability to promote cluster uniformity, our CDR addresses this limitation. In CC, an additional instance-wise contrastive loss is required to encourage instance uniformity, which unfortunately leads to the emergence of the class collision issue.
\end{itemize}

Moreover, our experimental analysis includes a comparison between the CDR approach and CC~\cite{li2021contrastive} within the same BYOL framework. We present the outcomes in Table. \ref{tab:6}, allowing for a comprehensive evaluation of performance. Notably, the results demonstrate that BYOL+CC exhibits reduced performance and unstable training, as CC does not effectively facilitate the promotion of uniform and standardized representations. Conversely, in the same circumstances, PIPCDR showcases remarkable enhancements over BYOL+CC, establishing its superiority in achieving significant improvements in performance.

\subsection{WCL and GCC}
To address the class collision issue, previous approaches such as WCL~\cite{zheng2021weakly} and GCC~\cite{zhong2021graph} employ graph-based methods to label neighboring samples as pseudo-positive examples. Subsequently, they utilize a supervised contrastive loss to encourage close proximity between the multiple pseudo-positive examples and the two data augmentations of a given example. While WCL forms the graph inside each batch, GCC utilizes a memory bank with moving averages for graph-oriented pseudo-labeling. The primary emphasis of GCC and WCL lies in effectively selecting positive instances from the memory bank or mini-batch to mitigate the issues related to the class collision. In this context, we classify the problem of class collision into the subsequent two scenarios:
\begin{itemize}
    \item[i)] \textit{Negative Class Collision Issue:} In the context of \ac{cl}, a notable challenge is the potential occurrence of a negative class collision issue. This issue arises when negative examples, which are intended to be distinct from positive examples, may not truly exhibit the desired dissimilarity.
    \item[ii)] \textit{Positive Class Collision Issue:}  The positive class collision issue is another pertinent concern in the realm of \ac{cl}. This issue, recently highlighted in the works of GCC and WCL, pertains to situations where positive examples, which should ideally represent similar instances, may not accurately capture the intended similarity.
\end{itemize}
Consequently, these approaches still face challenges regarding the collision issue within positive classes, given the possibility that the selected examples representing pseudo-positives might not accurately capture instances with high similarity. Moreover, they also encounter the negative class collision issue, as the inclusion of negative examples remains necessary for instance-wise \ac{cl}.

We delineate the distinctions between our proposed PIP and these existing approaches as follows:
\begin{itemize}
    \item While WCL and GCC choose instances from the dataset/memory bank/mini-batch, our approach PIP samples instances from the latent space, which may include examples not present in the dataset.
    \item While WCL and GCC pick neighboring instances as pseudo-positive instances in a graph, potentially lacking accuracy in representing positive instances, our approach PIP samples examples in the embedding space surrounding the instance, ensuring a more reliable representation of positive instances.
    \item While WCL and GCC  nonetheless depend on instance-wise contrastive loss, which can potentially lead to class collision issues, our approach employs a non-contrastive BYOL-method-based loss, allowing us to circumvent such issues and ensure more effective and reliable learning.
\end{itemize}
\subsection{Prototypical Contrastive Learning}
In this section, we outline the distinctions between our PIPCDR approach and PCL~\cite{li2020prototypical} specifically in terms of the employed losses. Below, we outline the dissimilarities between ProtoNCE loss and PIPCDR.
\begin{itemize}
    \item In contrast to PCL, our PIPCDR overcomes the class collision issue as it is built upon the BYOL framework, which eliminates the need for negative examples in \ac{rl}. In contrast, PCL utilizes instance-level contrastive loss, which requires a considerable number of negative instances and thereby gives rise to the class collision problem.
    \item CDR  offers a distinct conceptual approach compared to ProtoNCE employed in PCL. In PIPCDR, CDR aims to optimize the inter-cluster distance in order to encourage a uniform distribution across the space, while simultaneously minimizing the intra-cluster distance to enhance compactness within clusters. In contrast, ProtoNCE emphasizes enhancing compactness within clusters by minimizing the distance between instances and their respective clusters. Additionally, PIP further enhances the within-cluster compactness of PIPCDR.
    \item Pure CDR demonstrates strong performance in \ac{dc} tasks, as it inherently promotes the formation of a uniformly distributed space. In contrast, ProtoNCE relies on the assistance of another InfoNCE loss to achieve a similar outcome. The design of CDR allows for the maximization of inter-cluster distance, resulting in the formation of a uniformly distributed space. On the other hand, ProtoNCE faces challenges in avoiding collapse without the supplementary InfoNCE loss to facilitate the formation of such a space.
\end{itemize}

\bibliographystyle{IEEEtran}
\bibliography{bibtex/bib/IEEEabrv,bibtex/bib/IEEEexample}